\documentclass[journal,15pt]{IEEEtran}
\pdfoutput=1
\usepackage{times}
\usepackage{epsfig}
\usepackage{graphicx}
\usepackage{amsmath}
\usepackage{amssymb}
\usepackage{listings}
\usepackage{makecell}
\usepackage{booktabs}
\usepackage{multirow}
\usepackage{color}
\usepackage{comment}
\usepackage{icomma}
\usepackage{enumitem}
\usepackage{amsthm}
\usepackage[bf,font=footnotesize]{caption}
\usepackage{subfig}

\usepackage{dsfont}
\usepackage[table]{xcolor} 

\begin{document}
	\title{Multi-label Image Classification with Regional Latent Semantic Dependencies}
	
	\author{Junjie Zhang$^{*,1,3}$ \quad Qi Wu$^{*,2}$ \quad Chunhua Shen$^{2}$ \quad Jian Zhang$^{3}$ \quad Jianfeng Lu$^{1}$\\
		$^1$Nanjing University of Science and Technology, China \quad $^2$University of Adelaide, Australia \\
		$^3$University of Technology Sydney, Australia\\
		{\tt\small \{junjie.zhang@student.,jian.zhang@\}uts.edu.au  \quad \{qi.wu01,chunhua.shen\}@adelaide.edu.au} \\
		{\tt\small lujf@njust.edu.cn}
		\thanks{* Indicates equal contribution.}
	}
	
	\maketitle
	
	\begin{abstract}
		Deep convolution neural networks (CNN) have demonstrated advanced performance on single-label image classification, and various progress also have been made to apply CNN methods on multi-label image classification, which requires to annotate objects, attributes, scene categories etc. in a single shot. Recent state-of-the-art approaches to multi-label image classification exploit the label dependencies in an image, at global level, largely improving the labeling capacity. However, predicting small objects and visual concepts is still challenging due to the limited discrimination of the global visual features. In this paper, we propose a Regional Latent Semantic Dependencies model (RLSD) to address this problem. The utilized model includes a fully convolutional localization architecture to localize the regions that may contain multiple highly-dependent labels. The localized regions are further sent to the recurrent neural networks (RNN) to characterize the latent semantic dependencies at the regional level. Experimental results on several benchmark datasets show that our proposed model achieves the best performance compared to the state-of-the-art models, especially for predicting small objects occurred in the images. In addition, we set up an upper bound model (RLSD+ft-RPN) using bounding box coordinates during training, the experimental results also show that our RLSD can approach the upper bound without using the bounding-box annotations, which is more realistic in the real world.
	\end{abstract}
	
	
	\IEEEpeerreviewmaketitle
	
	\section{Introduction}
\label{intro}
\IEEEPARstart{L}{arge} scale images have become widely available due to the convenience of network access and the wide use of digital devices, which provide various opportunities for researchers to understand these images. As a traditional task, image classification has been comprehensively studied for decades, especially for the single-label classification problem \cite{krizhevsky2012imagenet,ILSVRC15}, various progresses have been made on it. However, in the real world, an image usually contains abundant semantic information, such as objects, attributes, actions and scenes etc. By assigning multiple labels to an image, we can transfer the vision information to language, which is more convenient to understand and useful for other vison applications such as image retrieval and semantic segmentation etc.

\begin{figure}[t]
	\centering
	\includegraphics[width=0.99\linewidth]{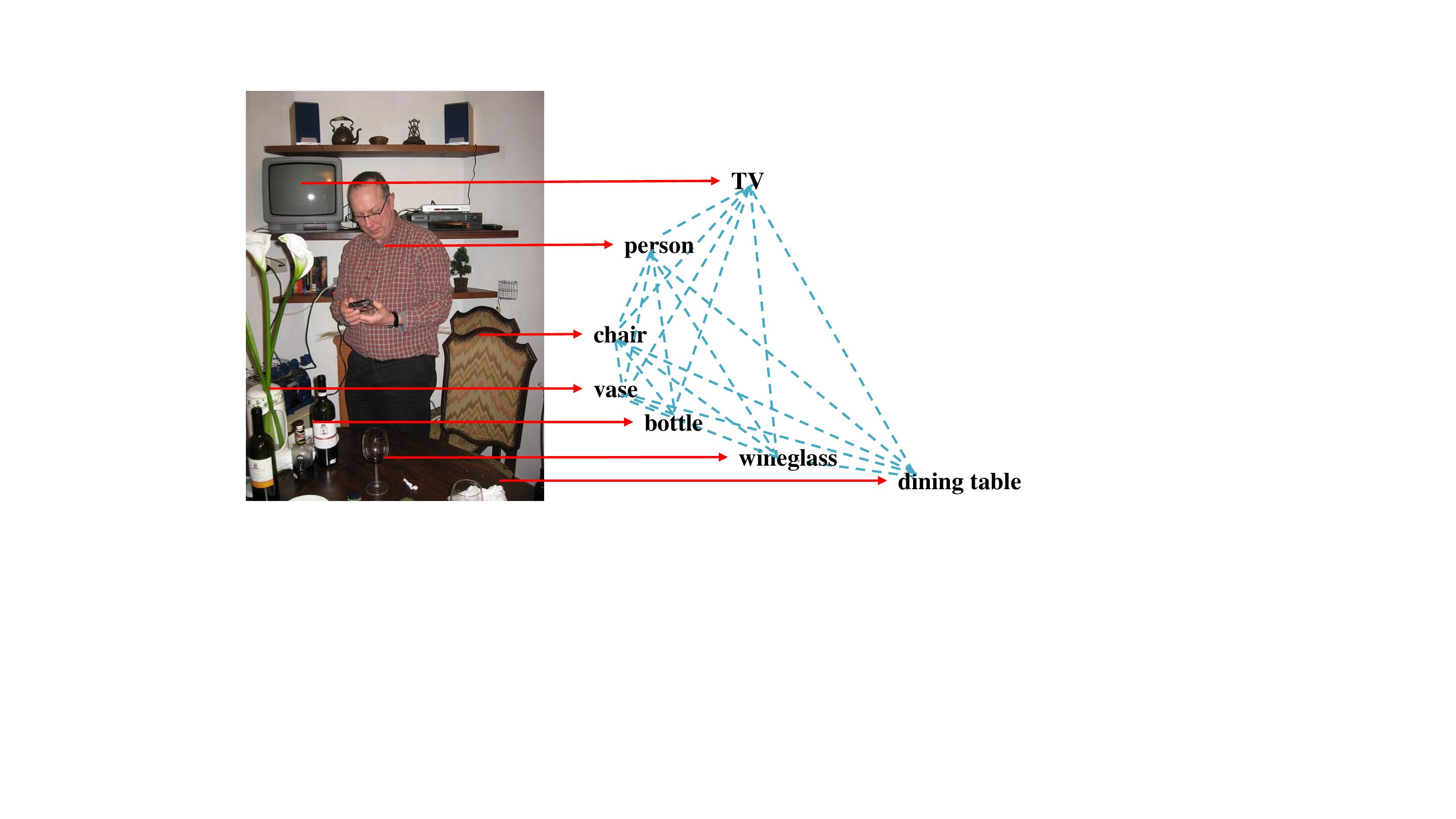}
	\vspace{-3pt}
	\caption{An example of multi-label image. Red arrows indicate the visual relevance between image content and labels, blue dot-lines indicate the semantic dependencies exist between labels.}
	\label{multilabel_example}
	\vspace{-15pt}
\end{figure}

The key issue behind this task is bridging the semantic gap existing between the image visual content and multiple labels. Fig. \ref{multilabel_example} shows an example of multi-label image. With the available of the large-scale dataset and the enrichment of the data annotations, multi-label image classification has drawn lots of attentions \cite{guillaumin2009tagprop,guo2011multi}. Inspired by the advanced performance of deep neural network, especially convolutional neural networks~(CNN) \cite{he2015deep,krizhevsky2012imagenet}, various efforts have been made to apply the neural network on the multi-label classification problem~\cite{wang2016cnn,single2multi}.

The most direct approach is to treat multi-label image classification problem as several separate single label classification problems, and to train the independent classifier for each label with a cross-entropy \cite{guillaumin2009tagprop} or ranking loss (such as WARP~\cite{gong2013deep}). Wei et al. \cite{single2multi} provide a regional solution allows predicting labels independently at the regional level. However, it is difficult for them to model the label dependencies between different labels. Intuitively, images with multiple labels usually contain strong correlations among the labels, for example, `ocean' and `ship' usually appear in the same image, while `ocean' and `cat' normally never occur together. To conveniently explore the label dependencies, the probabilistic graphical models (PGM) \cite{vanprobabilistic} are usually employed in the previous works \cite{li2014multi,tan2015learning}.

\begin{figure}[h]
	\centering
	\includegraphics[width=0.99\linewidth]{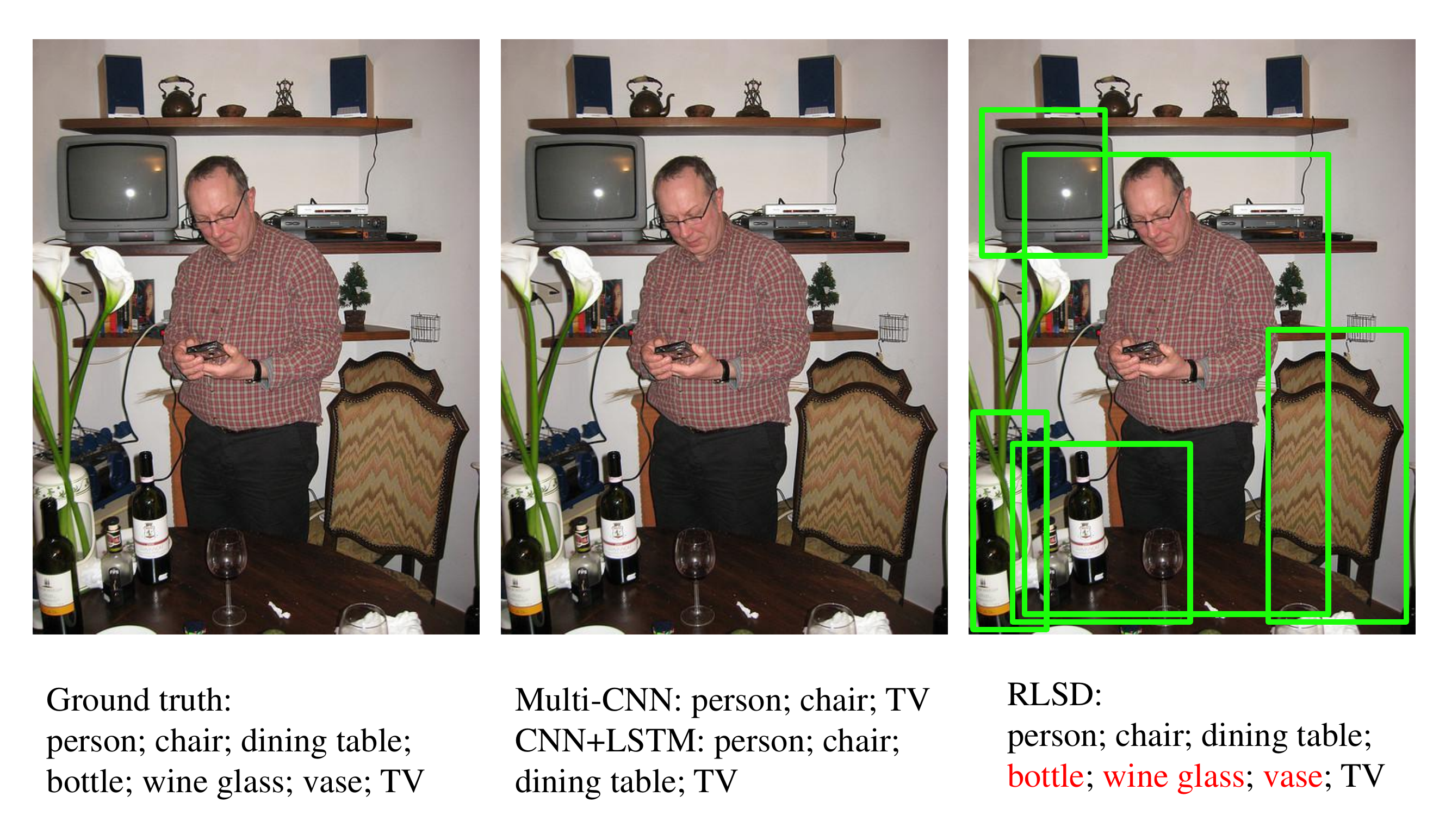}
	\vspace{-3pt}
	\caption{Example results of multi-label prediction from different models. The left is the ground-truth and the middle column shows the results from baseline models, Multi-CNN and CNN+LSTM. The right column displays the outputs of our proposed RLSD model, including predicted multiple labels and selected region proposals. Compare to the baseline methods, our model produces much richer predictions and is especially good at predicting small objects, such as `bottle', `wine glass' and `vase' etc.}
	\label{prediction_example}
	\vspace{-15pt}
\end{figure}

Most recently, Wang et al. \cite{wang2016cnn} have shown that the recurrent neural networks (RNN) \cite{hochreiter1997long,mikolov2010recurrent} can efficiently capture the high order label dependencies. They unify the CNN and RNN as one framework to exploit the label dependencies at the global level, largely improving the labeling ability. However, predicting small objects and attributes is still challenging for these works due to the limited discrimination of the global visual features. 

In this paper, our main contribution is that we propose a Regional Latent Semantic Dependencies (RLSD) model for the multi-label image classification, which effectively captures the latent semantic dependencies at the regional level. The proposed model combines the power of the region based features and the advantages of the RNN based label co-occurrence models, and achieves the best performance compared to the state-of-the-art multi-label classification models on several benchmark datasets, especially for predicting small objects and visual concepts. Fig.~\ref{prediction_example} shows an example output of our proposed RLSD model compared with the baselines models. We can see that the Multi-CNN and CNN+LSTM both fail to predict the `bottle', `vase' and `wine glass' in the image due to their small size, while our model efficiently predicts them along with other large objects.

\begin{figure*}[t]
	\centering
	\includegraphics[width=1\linewidth]{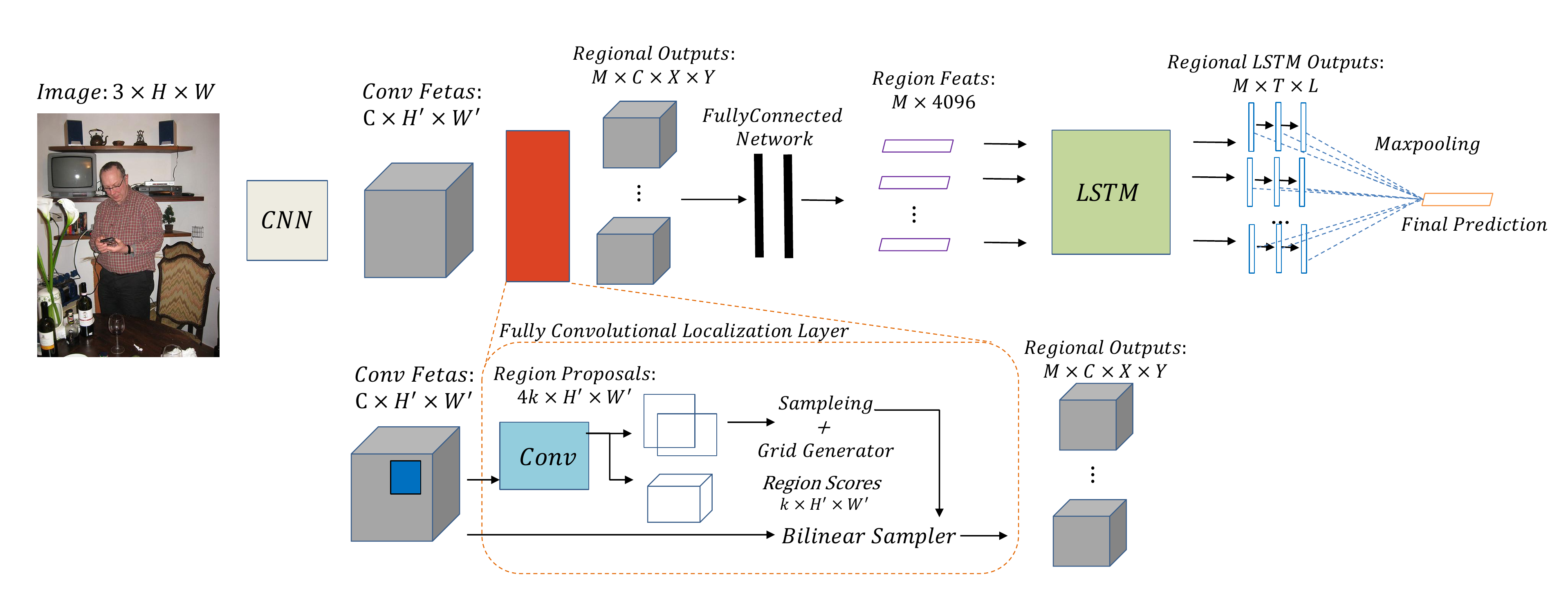}
	\caption{Our proposed Regional Latent Semantic Dependencies model. An input image is first processed through a CNN to extract convolutional features, which are further sent to an RPN-like fully convolutional localization layer. The localization layer localizes the regions in an image that potentially contain multiple highly-dependent labels. These regions are encoded with a fully-connected neural network and sent to the regional LSTM. Finally, a max-pooling operation is carried out to fuse all the regional outputs as the final prediction.}
	\label{framework}
\end{figure*}

The framework of the proposed model is shown in Fig.~\ref{framework}. An input image is first processed through a CNN to extract convolutional features, which are further sent to an RPN-like (Regional Proposal Network) localization layer. Different from the conventional RPN in the object detection framework (such as faster R-CNN \cite{ren2015faster}) which tries to predict the proposals with a single object inside, our localization layer is designed to localize the regions in an image that may contain multiple semantically dependent labels. These regions are encoded with a fully-connected neural network and further sent to a RNN, which captures the latent semantic dependencies at the regional level. The RNN unit sequentially outputs a multi-class prediction, based on the outputs of the localization layer and the outputs of previous recurrent neurons. Finally, a max-pooling operation is carried out to fuse all the regional outputs as the final prediction.

In addition, we also set up an upper bound model (RLSD+ft-RPN) by utilizing the object bounding-box coordinates for training. Our experimental results show that our model can approach this upper bound without involving additional bounding-box annotations, which is more realistic in the real world.

	\section{Related Works}
\label{rel_work}

Various works on multi-label image classification have been conducted over the past few years. The traditional bag-of-words (BoW) model \cite{chen2015contextualizing, chen2012hierarchical, dong2013subcategory, harzallah2009combining} relies on composing multiple modules, e.g. feature representation (SIFT \cite{lowe2004distinctive}, HOG \cite{dalal2005histograms}, LBP \cite{ojala1996comparative} etc.), classification (SVM \cite{chang2011libsvm}, random forests \cite{breiman2001random} ) and context modeling \cite{chen2015contextualizing, chen2012hierarchical, harzallah2009combining}.

Recent progresses in the image classification are made based on the powerful deep convolutional neural networks, which try to model the high-level abstractions of the visual data by using architectures composed of multiple non-linear transformations. Several approaches have been proposed to expand single label classification network \cite{he2015deep,krizhevsky2012imagenet,simonyan2014very} to the multi-label problem. Gong et al. \cite{gong2013deep} combine top-k ranking objectives with a CNN architecture to tackle this problem. By defining a weight function for pair-wised ranking labels, they minimize the loss function so that positive labels are ranked higher than negative ones. Wei et al. \cite{single2multi} provide a regional solution that allows predicting labels independently at the regional level. They use BING \cite{cheng2014bing} to generate object proposals and further send them into the CNN to compute multi-class scores. A max-pooling operation is applied to fuse the regional scores together as final classification results. We also use regional features and the max-pooling fusion. However, we consider the \textit{regional latent semantic dependencies}, which allow us to \textit{predict multiple labels jointly}.

There are also some works to solve the multi-label classification problem by designing a multi-modal representation, which bridges the semantic gap between the image and labels by learning the representation for the image visual content as well as labels. Canonical correlation analysis (CCA)~\cite{gong2014multi} and kernel canonical correlation analysis~(KCCA)~\cite{uricchio2016automatic} are usually carried out to build a latent semantic space to tackle the multi-label image annotation and retrieval problem. In addition, matrix completion \cite{cabral2011matrix} is also used to encode multi-label. These methods concentrate on digging labels' abundant semantic information, while neglect to explore the label dependencies among them.

To model the label dependencies, several approaches have been proposed. The probabilistic graphical models~\cite{vanprobabilistic} are usually employed in previous works \cite{bradley2010learning,ghamrawi2005collective,li2014multi,tan2015learning,zhang2010multi} to model the image feature-label joint distribution. There are several different graph structures to fulfill this purpose. For example, Chow-Liu tree \cite{chow1968approximating} is used to build a tree based on the labels' mutual dependency in~\cite{bradley2010learning}.  In~\cite{zhang2010multi}, joint probability is exploited by directed acyclic graph and chain rules. Conditional random field is used in \cite{ghamrawi2005collective,tan2015learning} and matrix completion is used in \cite{cabral2011matrix}. A limitation of the graph-based approaches is that the richer the label semantic information is, the more complex the graph can be, which causes high computational complexity and low efficiency. Moreover, all of the above methods only model the label dependencies at the global level.

Recurrent Neural Network (RNN) has been proved to be able to model the temporal dependency effectively in a sequence and it has been successfully applied in several sequence-to-sequence problems, such as image captioning \cite{densecap,karpathy2015deep,vinyals2015show,Wu_2016_CVPR}, visual question answering \cite{gao2015you, ren2015image, wu2015ask}, machine translation \cite{sutskever2014sequence}, speech recognition \cite{graves2013speech}, language modeling \cite{sundermeyer2012lstm}, etc. Most recently, Wang et al. \cite{wang2016cnn} have shown that the RNN \cite{hochreiter1997long,mikolov2010recurrent} can efficiently capture the high order label dependencies. They unify CNN and RNN into one framework to exploit the label dependencies at the global level, largely improving the labeling ability. We apply the RNN to capture the label dependencies as well. But different from \cite{wang2016cnn}, our regional latent semantic dependencies model considers the label dependencies at the \textit{regional level}, which allows us to \textit{predict small-size objects and visual concepts}. 

In summary, our proposed model is inspired by the previous works in image classification \cite{wang2016cnn,single2multi}, object detection \cite{ren2015faster} and image captioning \cite{densecap,vinyals2015show}. We propose to utilize the region proposal network, fully-connected recognition network and RNN together to extract image regions with abundant semantic information, and explore the latent semantic dependencies simultaneously. The following sections describe details of our proposed RLSD model.

\section{The RLSD Model}
\label{sec:model}

\paragraph{Framework Overview} 
The key characteristic of our proposed model is that it can capture the regional semantic label dependencies. The novelty lies in the fact that this is achieved by a localization architecture, followed by a few of LSTMs (Long-Short Term Memory). The purpose of the localization layer is to localize the regions that contain multiple highly-dependent labels, while the LSTMs are employed to characterize the latent semantic label dependencies in a sequential manner. A max-pooling operation is carried out to finally fuse all the regional outputs. Fig.~\ref{framework} shows the entire network of our proposed model.

In the following part, the localization layer is first introduced in Sec.~\ref{3.1} and the LSTM based label sequence prediction model is described in Sec.~\ref{3.2}. The final max-pooling operation and the loss function is summarized in Sec.~\ref{3.3}. The model initialization and some training details are given in Sec.~\ref{3.4}.

\subsection{Localizing Multi-label Regions}
\label{3.1}
To explore the image at the regional level, we need to generate the regions that potentially contain the multiple objects and visual concepts. Hence, the first component of our proposed model is to localize these regions. The conventional object proposal algorithms (such as Selective Search~\cite{UijlingsIJCV2013}, Objectness \cite{alexe2012measuring}, BING \cite{cheng2014bing} and MCG \cite{PABMM2015}, etc.) are ruled out since these methods only focus on predicting single object proposals, which means that a proposed region normally only contains one single object. Instead, Johnson~et al.~\cite{densecap} propose a fully convolutional neural network, extended from the Region Proposal Network (RPN)~\cite{ren2015faster}, to localize regions that can be described by a sentence, instead of a single label. Therefore, the proposed regions in~\cite{densecap} normally enjoy bigger label density, as well as the label complexity. Inspired by their work, we develop our approach of generating region proposals that are tailored for multi-label image classification.

\subsubsection{Convolutional Features as Input}
Since the convolutional layers of the CNN still preserve the spatial information of an image, which is necessary for us to explore the semantic dependencies at the regional level, we use them to extract image features. Specifically, we use the VGGNet~\cite{simonyan2014very} \footnote{We use VGGNet as basic CNN model for fair comparisons with state-of-the-art methods.} convolutional layer configuration, which consists of 13 convolutional layers with $3\times3$ kernel size and 5 max-pooling layers with $2\times2$ kernel size. The output of the last convolutional layer is used as image features. Given an input image with size $3\times H \times W$, the convolutional features will be $C \times H' \times W'$, where $H'=\lfloor\frac{H}{16}\rfloor$, $W'=\lfloor\frac{W}{16}\rfloor$, and $C=512$, same as in the VGGNet setting. The convolutional features are further sent to the localization layer to generate the region proposals that we are interested in.

\subsubsection{Fully Convolutional Localization Layer}
The input of the localization layer is the convolutional features extracted from the last step, while outputs are numbers of spatial regions of interest with a fixed-sized representation for each region.

\paragraph{Anchors and Regression}
By referring to \cite{densecap,ren2015faster}, we predict region proposals by regressing offsets from a set of generated anchors. Specifically, each point inside the convolutional feature map is projected back into the original image ($H \times W$), and futher used as center to generate $k$ different aspect ratio anchor boxes. Each anchor box is sent into a fully convolutional network to produce the predicted box scalars and a confidence score. The fully convolutional network consists of 256 convolutional filters with $3 \times 3$ kernel size, a ReLU layer and a final convolutional layer with $(4+1) \times k$ filters, where 4 stands for the number of the box scalars, and 1 stands for the confidence score. We set $k=12$ in our proposed models. As for the bounding-box regression from the anchors to the region proposals, we refer to ~\cite{ren2015faster} for the parameterization. We apply the log-space scaling transformations on the anchor box, which means given an anchor box's parameters $(a_x,a_y,a_w,a_h)$, where $(a_x,a_y)$ is the center of the anchor box, and $a_w, a_h$ stands for the width and height of the anchor box respectively, we generate the region coordinates $b = (b_x,b_y,b_w,b_h)$ by following folumas:

\begin{eqnarray}
b_x = a_x + t_{x}a_w\qquad b_y = a_y + t_{y}a_h\\
b_w = a_wexp(t_{w})\qquad b_h = a_hexp(t_{h})
\end{eqnarray}

The scalars $t_x,t_y,t_w,t_h$ are predicted by our model. Smooth $L_1$ norm is employed as the loss function to regress the region location \footnote{This loss is only used when we fine-tune the localization layer in the upper-bound model RLSD+ft-RPN.}. Given the ground-truth coordinates $g = (g_x,g_y,g_w,g_h)$, the loss function is defined as:

\begin{eqnarray}
L(b,g) = \sum_{i\in{x,y,w,h}} Smooth_{L_1}(b_i,v_i)
\end{eqnarray}

where

\begin{equation}  
Smooth_{L_1} = \left\{  
\begin{array}{lr}  
0.5x^2\qquad & if |x|<1\\  
|x|-0.5\qquad & otherwise\\
\end{array}  
\right.  
\end{equation} 

\paragraph{Box Sampling and Bilinear Interpolation}

A sampling mechanism is employed here to subsample the generated region proposals, since it is expensive to send all the proposals to the further LSTM-based label generation step. By referring to \cite{densecap,ren2015faster}, a $M=256$ size minibatch is sampled. The regions with top $M/2$  highest confidence score are considered as the positive samples, and the lowest $M/2$ regions are negative. \footnote{In RLSD+ft-RPN, A box with IoU (Intersection of Union) which is larger than 0.7 ratio of the ground-truth region is considered as the positive sample, and less than 0.3 as the negative.} We also restricted that at most of the half of boxes in one minibatch are positive samples and the other half are negative. During the test stage, non-maximum suppression is used to select the top $M$ highest ranked proposals \footnote{In practical, the entire image is also added into the proposals, since some labels are related to the whole image.}.

To ensure that the region proposal features can be accepted by the fully-connected layer and gradients can be back propagated to both the input features and box coordinates, the bilinear interpolation is used to replace the ROI pooling layer in \cite{ren2015faster}. We refer to the bilinear sampling operation in \cite{densecap}, which results in $M\times C\times X\times Y$ feature maps for the top $M$ region proposals, where $C=512$ is the VGGNet convolutional feature map size, and $X$, $Y$ are the bilinear sampling grid size. In our case, we set $X=Y=7$, referring to \cite{jaderberg2015spatial,densecap}. 

\subsubsection{Encoded by a Fully-Connected Network}

After the regional features as $M\times C\times X\times Y$ are obtained, they are sent to a fully-connected network, which is formed by two 4096-d fully-connected layers and regularized by dropout. Features from each region are flattened into a vector and passed through this fully-connected network. Thus, each proposal region is encoded as a feature vector $v$ with 4096 dimensions. All the regions' fully-connected features form a minibatch $V=[v_1,v_2,...v_i,...v_M]$ with the size of $M \times 4096$, where $i$ indicates the $i^{th}$ region proposal.

\begin{figure}[t!]
	\centering
	\resizebox{\linewidth}{!}{
		\begin{tabular}{c c}
			\includegraphics[height=3.5cm]{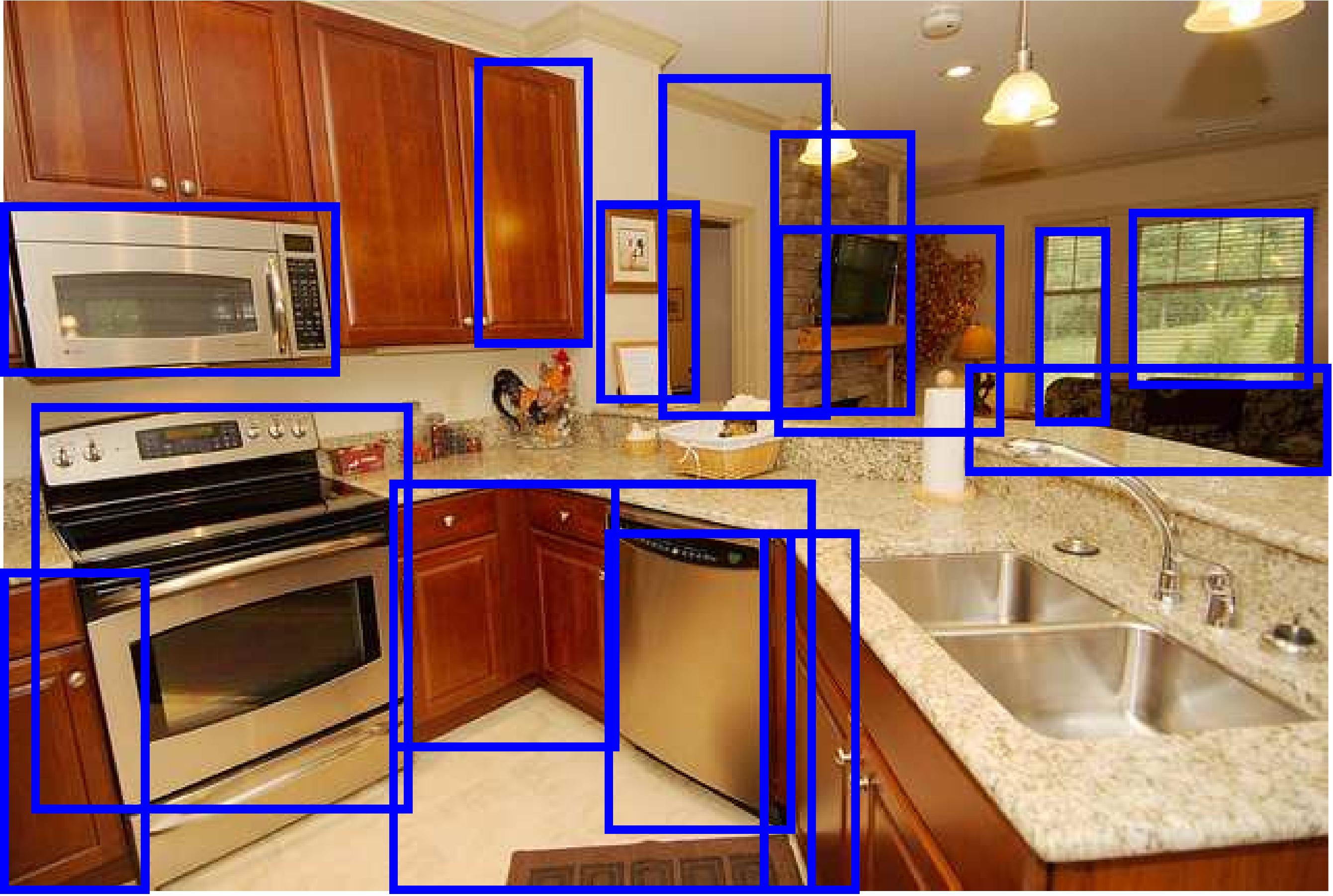}&
			\includegraphics[height=3.5cm]{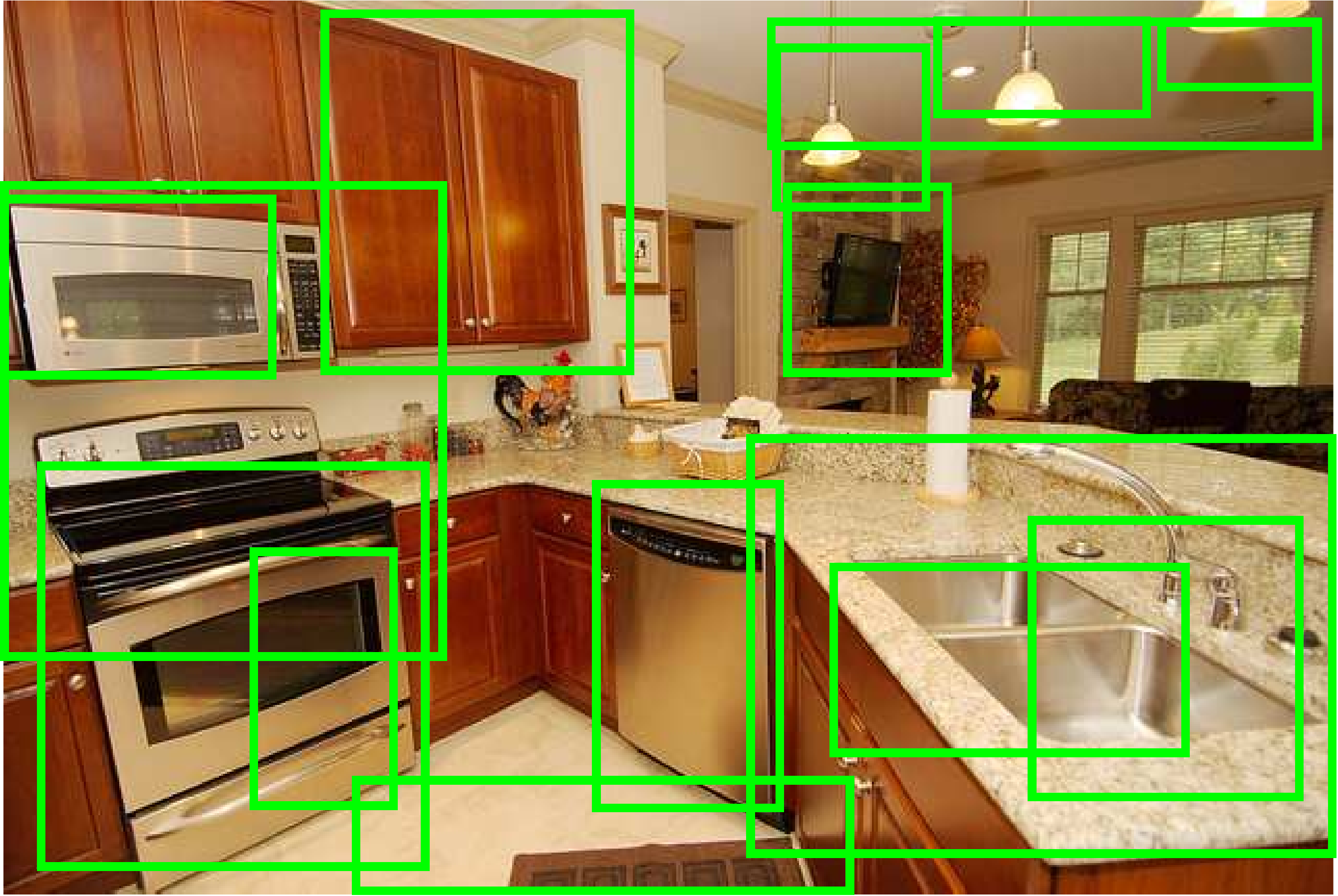} \\
			\includegraphics[height=4cm]{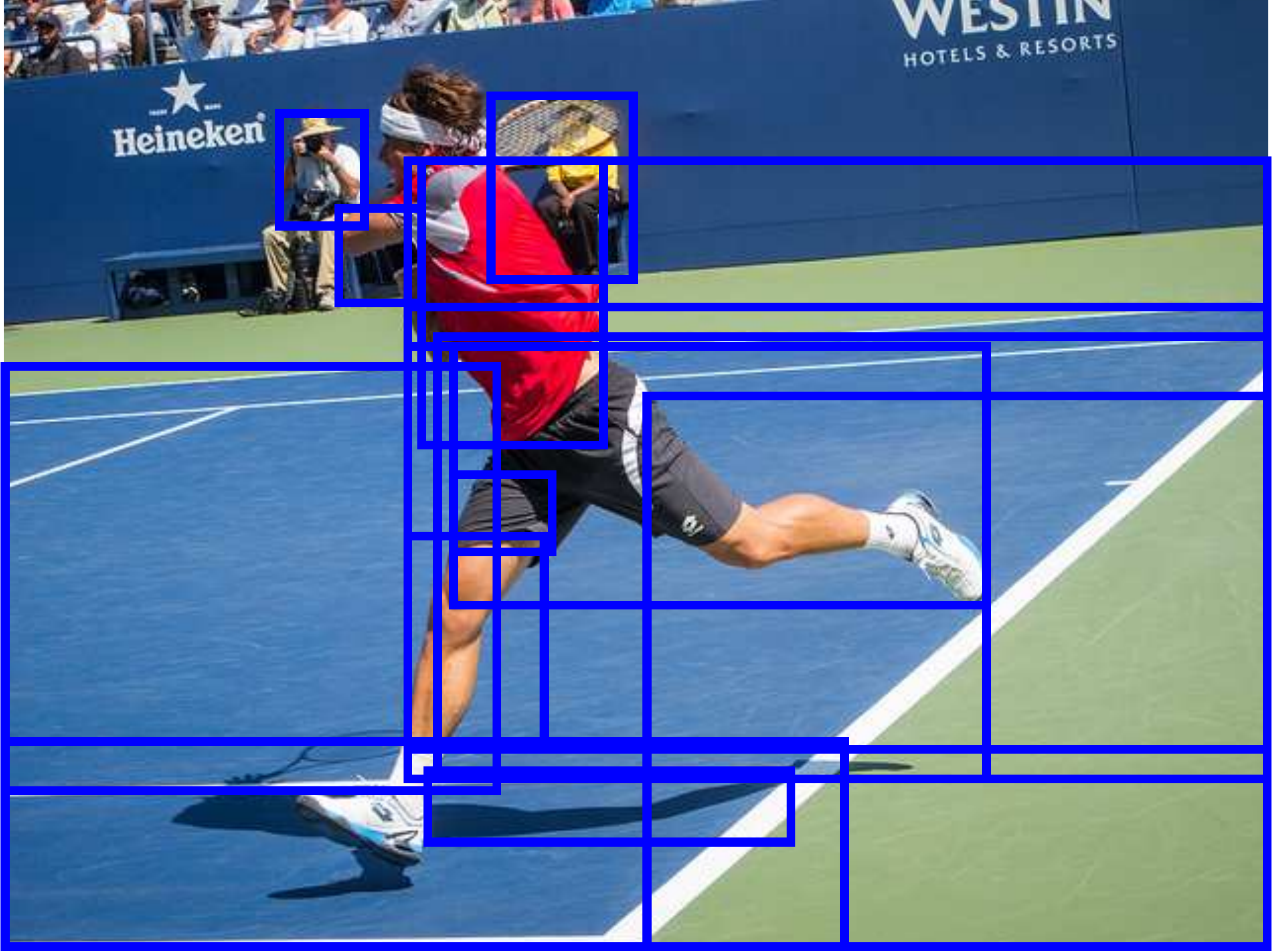} & \includegraphics[height=4cm]{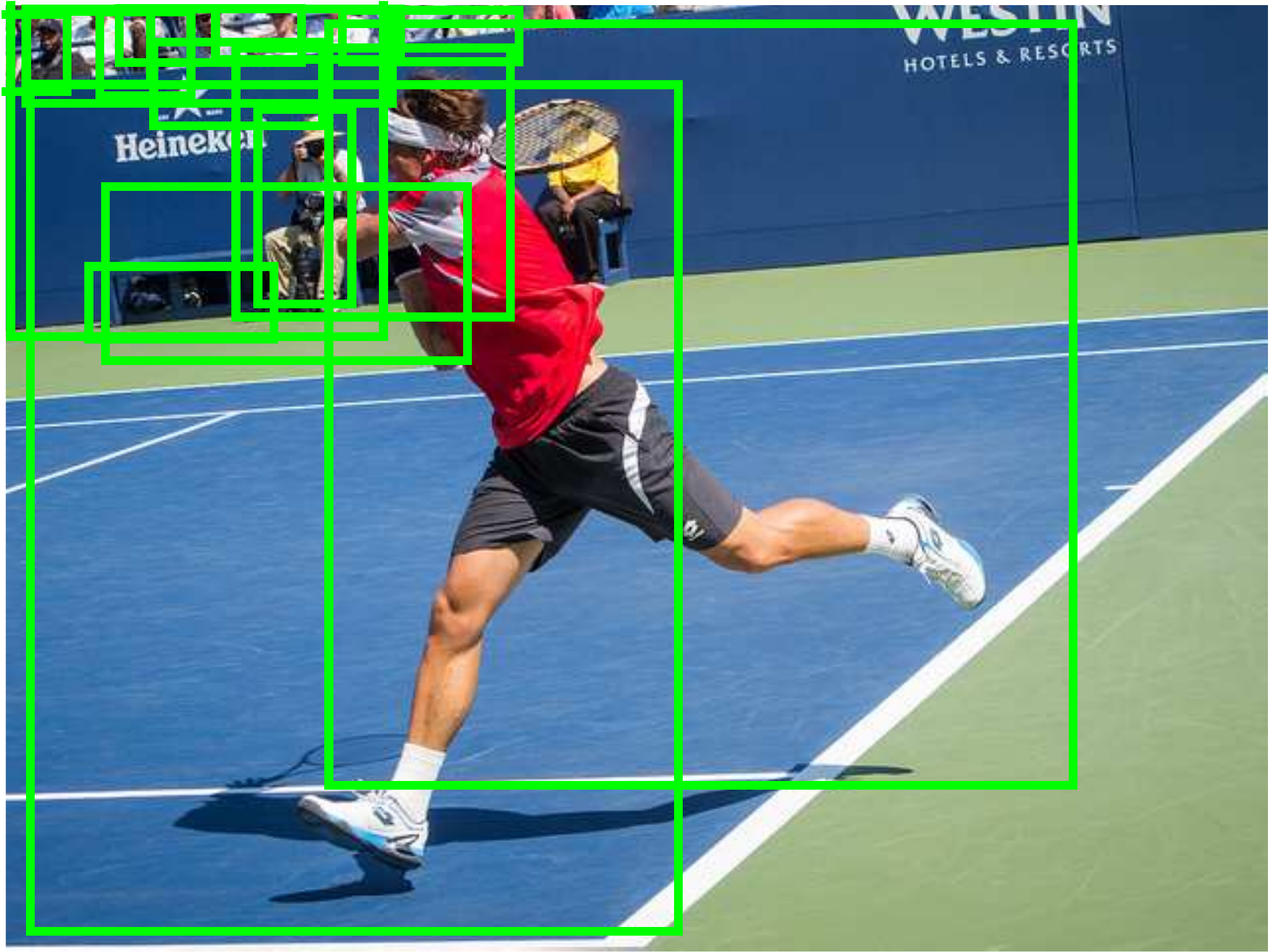} \\
			\includegraphics[height=4cm]{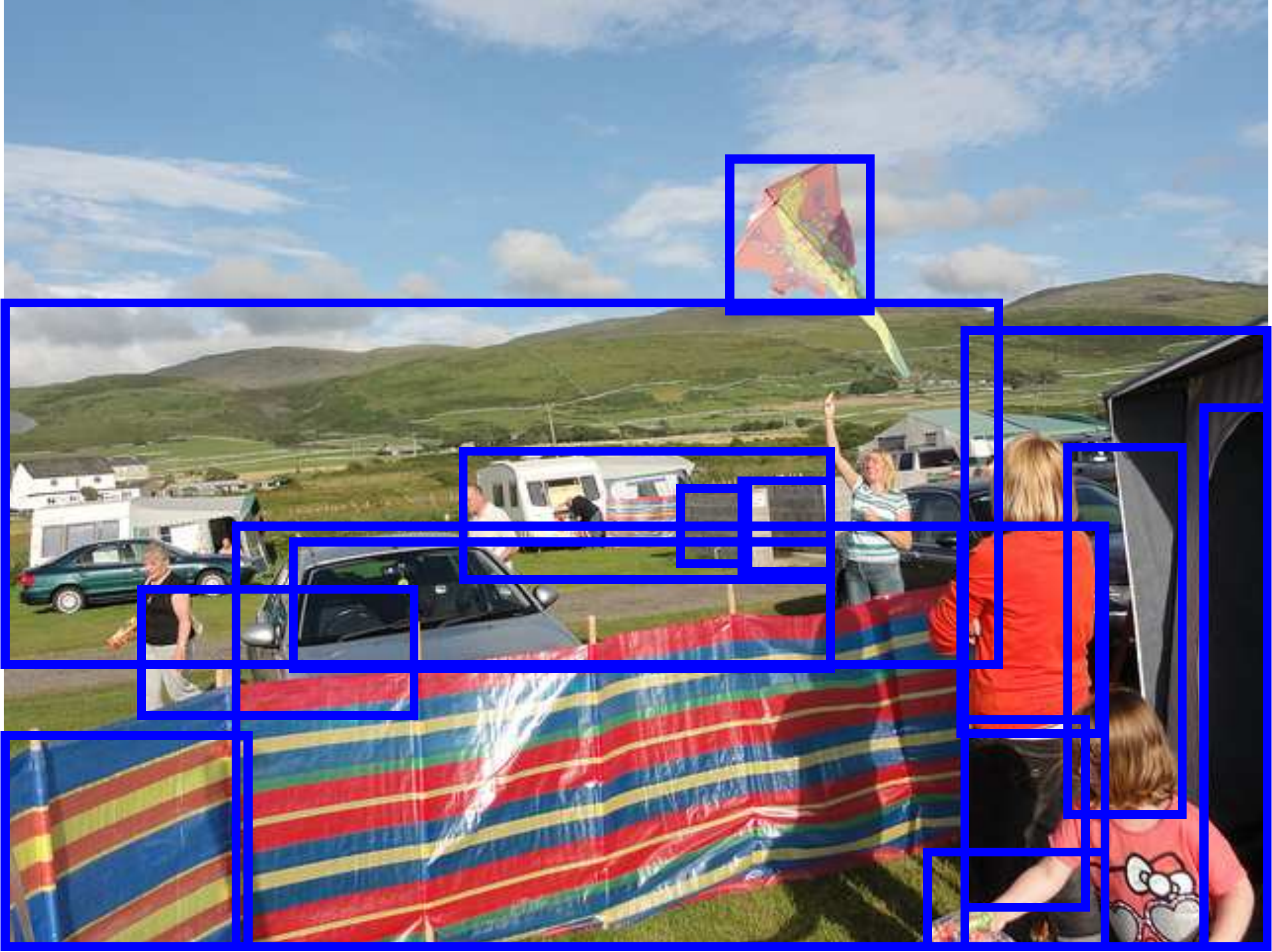} & \includegraphics[height=4cm]{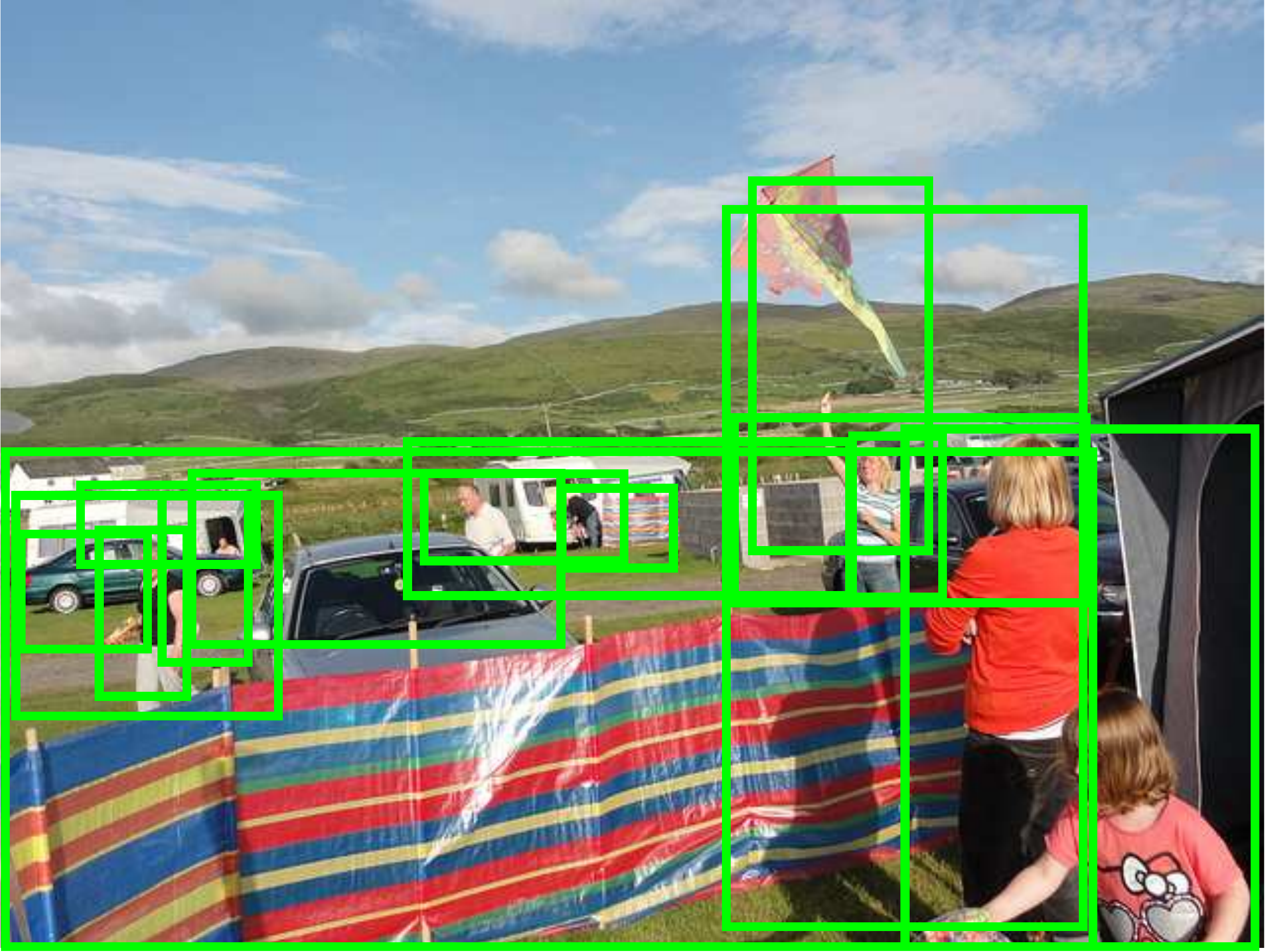} \\
			
	\end{tabular}}
	\caption{The comparison results between the Top-15 regions generated by MCG \cite{PABMM2015} (left) and our localization layer (right). Some of our generated regions contain multiple objects, for example, the generated regions contain the object of `oven/microwave/kitchenwares', `person/tennis racket', and `person/kite/car' all together.}
	\label{fig:region_proposals}
\end{figure}

Fig.~\ref{fig:region_proposals} shows some examples of the comparison results between our localization layer proposed regions and the MCG \cite{PABMM2015} produced object proposals. The bounding boxes generated by our model are normally bigger, and some of them contain multiple objects. Therefore, our model not only can explore the sufficient label dependencies, but also can outperform the current methods in predicting small objects and visual concepts. To show the effectiveness of the localization layer, we set a baseline model that uses the MCG \cite{PABMM2015} to replace our multi-label region localization layer for the further multi-label classification. More details can be found in the Sec.~\ref{exp}.

\subsection{An LSTM-based Multi-Label Generator}
\label{3.2}

To capture the latent semantic dependencies existed in those regions, we employ LSTMs to generate the sequence of label probability distributions on each single region. 

\begin{figure}[h]	
	\centering
	\includegraphics[width=1\linewidth]{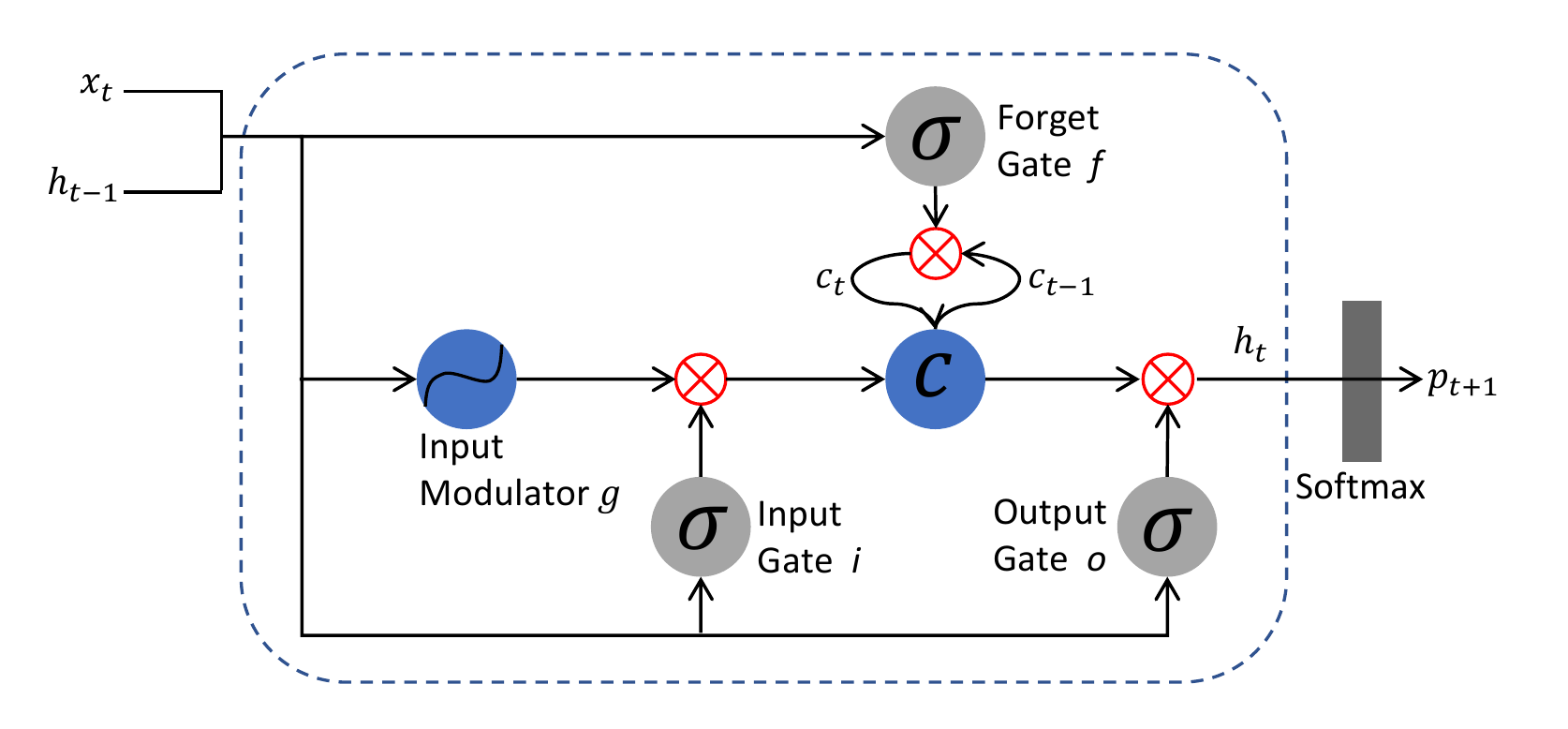}
	\caption{The structure of LSTM.}
	\label{LSTM}
\end{figure}

The LSTM is a memory cell which encodes the knowledge at every time step for what inputs that have been observed up to this step. Fig. \ref{LSTM} shows the basic structure of LSTM. We follow the model used in \cite{hochreiter1997long}. Letting $\sigma$ be the sigmoid nonlinearity, the LSTM updates for time step $t$ given inputs $x_t$, $h_{t-1}$, $c_{t-1}$ are:
\begin{eqnarray}
i_t &=& \sigma(W_{xi}x_t+W_{hi}h_{t-1}+b_i) \label{eq_start}\\
f_t &=& \sigma(W_{xf}x_t+W_{hf}h_{t-1}+b_f) \\
o_t &=& \sigma(W_{xo}x_t+W_{ho}h_{t-1}+b_o) \\
g_t &=& \tanh(W_{xc}x_t+W_{hc}h_{t-1}+b_c) \\
c_t &=& f_t\odot c_{t-1}+i_t\odot g_t \\
h_t &=& o_t \odot \tanh(c_t) \\
p_t &=& {\rm softmax}(h_t) \label{eq_end}
\vspace{-0.1cm}
\end{eqnarray}
Here, $i_t, f_t, c_t, o_t$ are the input, forget, memory, output state of the LSTM. The various $W$ matrices are trained parameters and $\odot$ represents the product with a gate value. $h_t$ is the hidden state at time step $t$ and is fed to a Softmax, which will produce a probability distribution $p_t$ over all labels.

Given a region feature vector $v$, we set $x_0=W_{ev}v$, where $W_{ev}$ is the learnable region features embedding weights. Following the equations (\ref{eq_start}) to (\ref{eq_end}), it gives us an initial hidden state $h_0$ which can be used in the next time step. From $t=1$ to $t=T$, we set $x_t=W_{es}S_t$ and the hidden state $h_{t-1}$ is given by the previous step, where $W_{es}$ is the learnable label embedding weights. $T$ is the numbers of label in a region and $S_t$ is the input label at time step $t$ \footnote{In practical, a label is represented as a one-hot vector, where 1 indicates the label exists, and 0 elsewhere.}. Practically, in our RLSD model, since only the global multi-label ground-truth are provided and no regional ground-truth can be used in the training stage (as well as in the testing), we call the $S_t$ as a latent label, which can be obtained by the following equation:
\begin{equation}
S_t^i=\mathds{1}\{p_{t-1}^i=\max(p_{t-1})\}
\end{equation}
where $\mathds{1}$ is an indicator function and $S_t$ is a one-hot vector that index $i=1$, and 0 elsewhere. $i$ is the index of the maximum value of the probability distribution $p_{t-1}$ over all the labels, which is computed by the LSTM feed-forward process at the previous $t-1$ time step. After all the labels in a region are predicted, an `END' label is added to finalize the prediction.

Giving all the $M$ region features in a minibatch (all the regions in a minibatch come from the same image) into the LSTM model, we gather the prediction $p_{tm}$ at each time step $t$, on each region $m$, to form a matrix with the shape $M \times T \times L$, where $L$ is the label size of a dataset. If a region label length is less than $T$, we will pad $0$.

\subsection{Max-pooling and Loss Function}
\label{3.3}
To suppress the possibly noisy prediction on some region proposals or at certain time steps, a cross region and time max-pooling is carried out to fuse the outputs into one integrative prediction. Suppose $p_{tm}$ is the output prediction of region $m$ at time step $t$ and $p_{tm}^{(j)} (j = 1,..., L)$ is the $j^{th}$ component of $p_{tm}$. The max-pooling in the fusion layer can be formulated as:

\begin{equation}
p^{(j)} = \max(p_{11}^{(j)},...,p_{1m}^{(j)},p_{21}^{(j)},...,p_{TM}^{(j)})
\end{equation}
where $p^{(j)}$ can be considered as the predicted value for the $j^{th}$ category of the given image.

The max-pooling fusion is a crucial step for the proposed RLSD model to be robust to the noise. The output of the fusion layer is fed into a multi-way softmax layer with the squared loss as the cost function, which is defined as:

\begin{equation}
J=\frac{1}{N}\sum_{i=1}^{N}\sum_{l=1}^L(p_{il}-\hat{p}_{il})^2
\end{equation}
where $\hat{p}_{i}=y_i/||y_i||_1$ is the ground-truth probability vector of $i^{th}$ image and $p_{i}$ is the predictive probability vector of $i^{th}$ image. $N$ is the number of images.\\

\begin{figure*}[h]	
	\centering
	\includegraphics[width=1\linewidth]{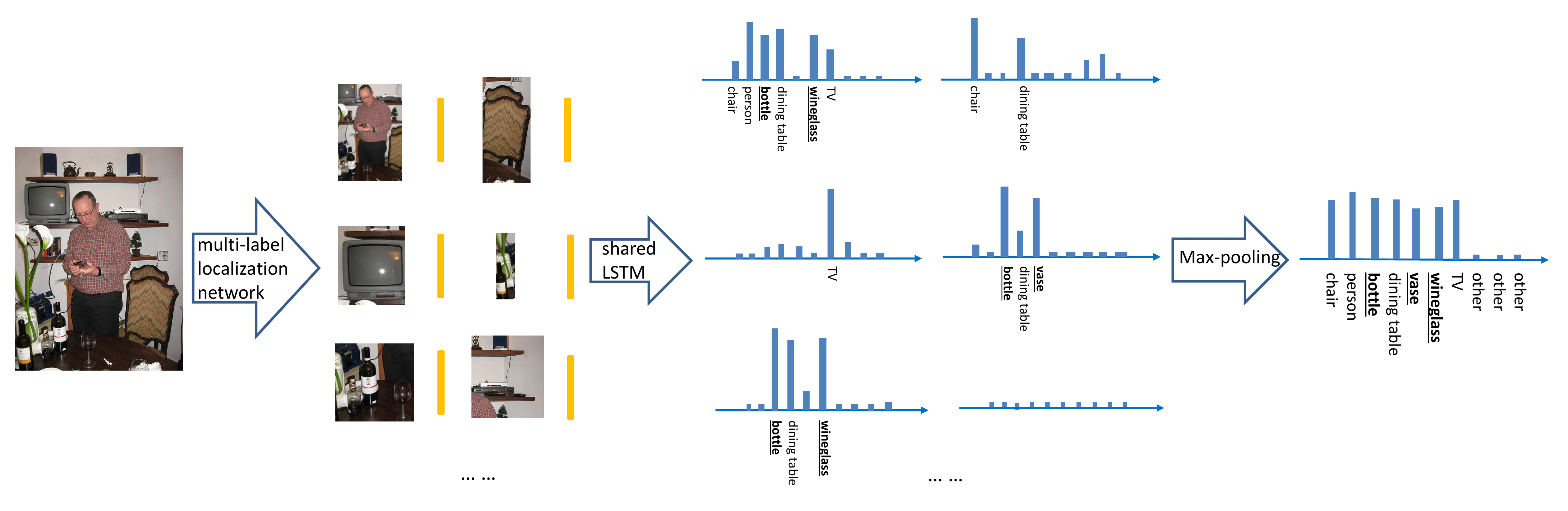}
	\caption{An illustration of proposed RLSD model for test image. The potential multi-label regions of test image are generated by localization layer, and further used to extract features and input to shared LSTM. As we can see, the small sized objects like `wine glass', `bottle' and `vase' etc. can be included inside the regions due to our multi-label localization network. The test is also performed in an end-to-end fashion.}
	\label{test_demo}
\end{figure*}

Fig .\ref{test_demo} shows the illustration of proposed RLSD model for test image. The potential multi-label regions of test image are generated by localization layer, and further used to extract features and input to shared LSTM. As we can see, the small sized objects like `wine glass', `bottle' and `vase' etc. can be included inside the regions due to our multi-label localization network. The test is also performed in an end-to-end fashion.

\subsection{Initialization and Pre-Training}
\label{3.4}
Our model is capable of training end-to-end from scratch, but a proper initialization and pre-training mechanism is important to achieve a promising performance.

\paragraph{Localization Layer Pre-Training} The localization layer is pre-trained on the Visual Genome region caption dataset~\cite{krishnavisualgenome}. Different from other object detection datasets, each region in the image of this dataset normally contains several object and visual concepts, which is very suitable for our multi-label region localization task.

\paragraph{LSTM Pre-training}
In the training stage, the LSTM is first pre-trained on the global image without region proposals, where every time step has the global image label as ground-truth to compute loss \footnote{This global model is also implemented and compared in the experimental part, known as CNN-LSTM}. Then the pre-trained LSTM is used as the initialization of the regional LSTM in our proposed RLSD model. 
We find this initialization process is important for model to fast converge.

	\section{Experiments}
\label{exp}
In this section, we present our experimental results and analysis to show the effectiveness of our proposed RLSD model for multi-label image classification problem. We evaluate the proposed model on three benchmark datasets: VOC PASCAL 2007 \cite{Everingham15}, Microsoft COCO \cite{lin2014microsoft} and NUS-WIDE \cite{nus-wide-civr09}. By comparing with several state-of-the-art models and baseline models, we show that our proposed RLSD model achieves the best performance. We further analyze the precision-recall along with the bounding-box size to show our model is especially good at predicting small objects.

\begin{table*}[t!]
	\centering
	\scriptsize
	\resizebox{\linewidth}{!}{
		\rowcolors{2}{gray!25}{} 
		\begin{tabular}{cp{0.3cm}p{0.3cm}p{0.3cm}p{0.3cm}p{0.3cm}p{0.3cm}p{0.3cm}p{0.3cm}p{0.3cm}p{0.3cm}p{0.3cm}p{0.3cm}p{0.3cm}p{0.3cm}p{0.3cm}p{0.3cm}p{0.3cm}p{0.3cm}p{0.3cm}p{0.3cm}|p{0.3cm}}
			\Xhline{2\arrayrulewidth}
			& plane         & bike          & bird          & boat          & bottle        & bus           & car           & cat           & chair         & cow           & table         & dog           & horse         & motor         & person        & plant         & sheep         & sofa          & train         & tv   & mAP           \\ \Xhline{2\arrayrulewidth}
			INRIA \cite{harzallah2009combining}           & 77.2          & 69.3          & 56.2          & 66.6          & 45.5          & 68.1          & 83.4          & 53.6          & 58.3          & 51.1          & 62.2          & 45.2          & 78.4          & 69.7          & 86.1          & 52.4          & 54.4          & 54.3          & 75.8          & 62.1 & 63.5          \\
			FV \cite{perronnin2010improving}          & 75.7          & 64.8          & 52.8          & 70.6          & 30.0          & 64.1          & 77.5          & 55.5          & 55.6          & 41.8          & 56.3          & 41.7          & 76.3          & 64.4          & 82.7          & 28.3          & 39.7          & 56.6          & 79.7          & 51.5 & 58.3          \\
			CNN-SVM \cite{sharif2014cnn}         & 88.5          & 81.0          & 83.5          & 82.0          & 42.0          & 72.5          & 85.3          & 81.6          & 59.9          & 58.5          & 66.5          & 77.8          & 81.8          & 78.8          & 90.2          & 54.8          & 71.1          & 62.6          & 87.4          & 71.8 & 73.9          \\
			I-FT \cite{single2multi}           & 91.4          & 84.7          & 87.5          & 81.8          & 40.2          & 73.0          & 86.4          & 84.8          & 51.8          & 63.9          & 67.9          & 82.7          & 84.0          & 76.9          & 90.4          & 51.5          & 79.9          & 54.3          & 89.5          & 65.8 & 74.5          \\
			HCP-1000C \cite{single2multi}      & 95.1          & 90.1          & 92.8          & 89.9          & 51.5          & 80.0          & 91.7          & 91.6          & 57.7          & 77.8          & 70.9          & 89.3          & 89.3          & 85.2          & 93.0          & 64.0          & 85.7          & 62.7          & 94.4          & 78.3 & 81.5          \\
			HCP-2000C \cite{single2multi}      & 96.0          & 92.1          & 93.7          & 93.4          & 58.7          & 84.0          & 93.4          & 92.0          & 62.8          & 89.1          & 76.3          & 91.4          & 95.0          & 87.8          & 93.1          & 69.9          & 90.3          & 68.0          & 96.8          & 80.6 & 85.2          \\
			CNN-RNN \cite{wang2016cnn}        & 96.7          & 83.1          & \textbf{94.2} & 92.8          & 61.2          & 82.1          & 89.1          & 94.2          & 64.2          & 83.6          & 70.0          & 92.4          & 91.7          & 84.2          & 93.7          & 59.8          & 93.2          & 75.3          & \textbf{99.7} & 78.6 & 84.0          \\ \hline
			Multi-CNN       & 95.8          & 88.7          & 93.2          & 91.1          & 51.5          & 82.1          & 89.5          & 91.8          & 68.3          & 80.2          & 75.7          & 91.4          & 92.6          & 88.7          & 92.8          & 61.2          & 82.9          & 68.0          & 96.2          & 78.0 & 83.0          \\
			CNN+LSTM        & \textbf{96.8} & 89.7          & 93.3          & 90.4          & 54.6          & 85.6          & 89.0          & 92.3          & 68.9          & 80.9          & \textbf{76.8} & 90.5          & 93.6          & 89.5          & 93.0          & 60.3          & 84.3          & 67.4          & 96.2          & 76.7 & 83.5          \\
			MCG-CNN+LSTM        & 94.3        & 90.4          & 91.4          & 91.4          & 64.9          & 90.3          & 92.9          & 94.3          & 71.9          & 83.7          & 55.6          & 92.5          & 94.5          & 90.0          & 96.8          & 60.1          & 90.5          & 71.3          & 96.3          & 84.7 & 84.9          \\ \hline
			RLSD & 95.3          & 92.4          & 91.2          & 92.1          & \textbf{71.9} & 91.1          & 93.3          & 94.8          & \textbf{74.9} & 86.1          & 70.4          & 93.3          & 95.6          & 89.7          & \textbf{98.0} & 66.8          & 89.4          & \textbf{75.7} & 96.6          & 85.9 & 87.3          \\
			RLSD+ft-RPN  & 96.4          & \textbf{92.7} & 93.8          & \textbf{94.1} & 71.2          & \textbf{92.5} & \textbf{94.2} & \textbf{95.7} & 74.3          & \textbf{90.0} & 74.2          & \textbf{95.4} & \textbf{96.2} & \textbf{92.1} & 97.9          & \textbf{66.9} & \textbf{93.5} & 73.7          & 97.5          & \textbf{87.6} & \textbf{88.5} \\ \Xhline{2\arrayrulewidth}
	\end{tabular}}
	\caption{Comparisons of classification results (AP in \%)  on the PASCAL VOC 2007 dataset.}
	\vspace{-10pt}
	\label{voc2007}
\end{table*}

\subsection{Implementation Details}
\label{details}
As mentioned in the Sec.~\ref{3.1}, the VGGNet \cite{simonyan2014very} is used to initialize the convolutional layers for the localization network. The embedding size of the image region feature (dimension of $W_{ev}$) and the label embedding size is 64 (dimension of $W_{es}$). We only use the one-layer LSTM, while the LSTM memory cell size is 512. Stochastic gradient descent (SGD) is used for optimization and we employ learning rate $10^{\text{-}5}$, momentum 0.9 and drop rate 0.5.

\paragraph{Data preprocessing}
For VOC 2007 \cite{Everingham15} and MS-COCO dataset \cite{lin2014microsoft}, the ground-truth of bounding-boxes for the single objects are provided. Therefore, we can use the bounding-box coordinates to fine tune the localization layer in our model. By using this additional information, the fine-tuned model can extract the regions more accurately. Therefore, it can be viewed as an upper bound model to verify the RLSD's generality. We call this upper bound model RLSD+ft-RPN (RLSD with fine-tuned RPN). The data pre-processing procedures are: given an image with several single-object bounding-boxes, we first compute their centers. Since the relevant objects usually appear close to each other, Euclidean distance based hierarchical clustering is applied on these bounding-box centers to generate several clusters. Bounding-boxes which belong to the same cluster are merged as one novel region. These novel regions are further used as the ground-truth to fine-tune the localization layer. Please note that only the bounding-box coordinates are used in this pre-processing, there are no label annotations involved.

\begin{figure}[t!]
	\centering
	\resizebox{\linewidth}{!}{
		\begin{tabular}{c c}
			\includegraphics[height=4cm]{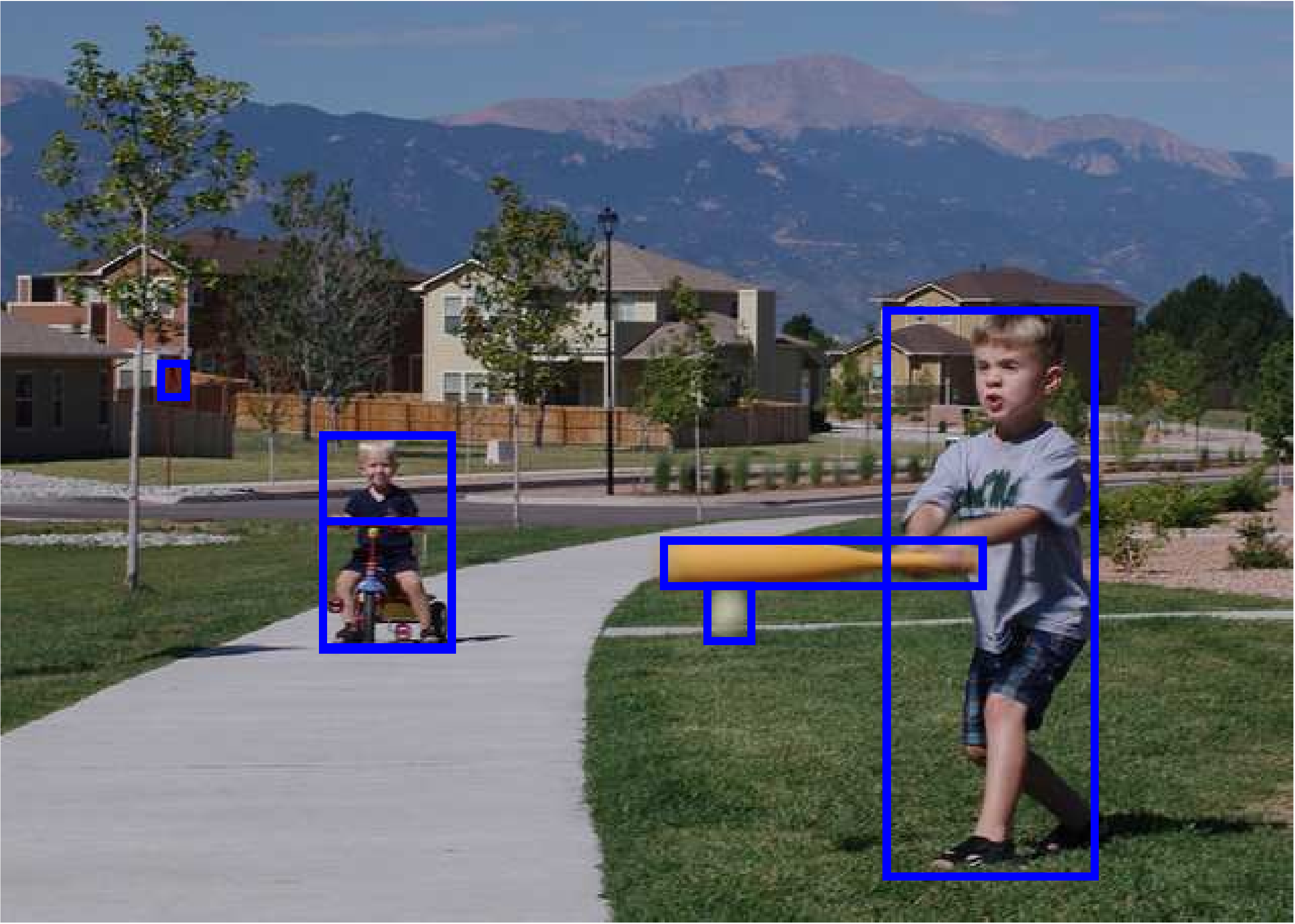}&
			\includegraphics[height=4cm]{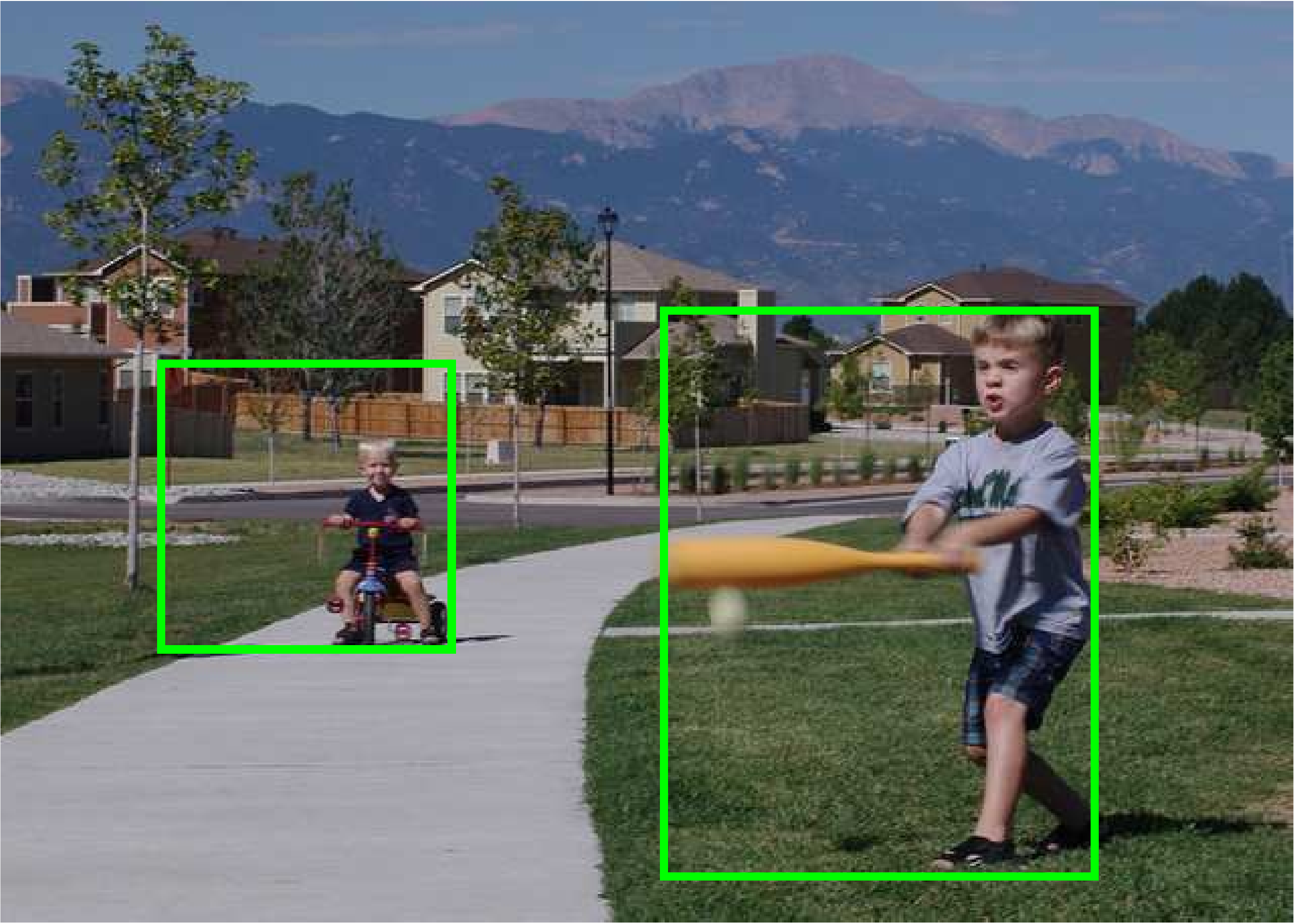} \\
			\includegraphics[height=4cm]{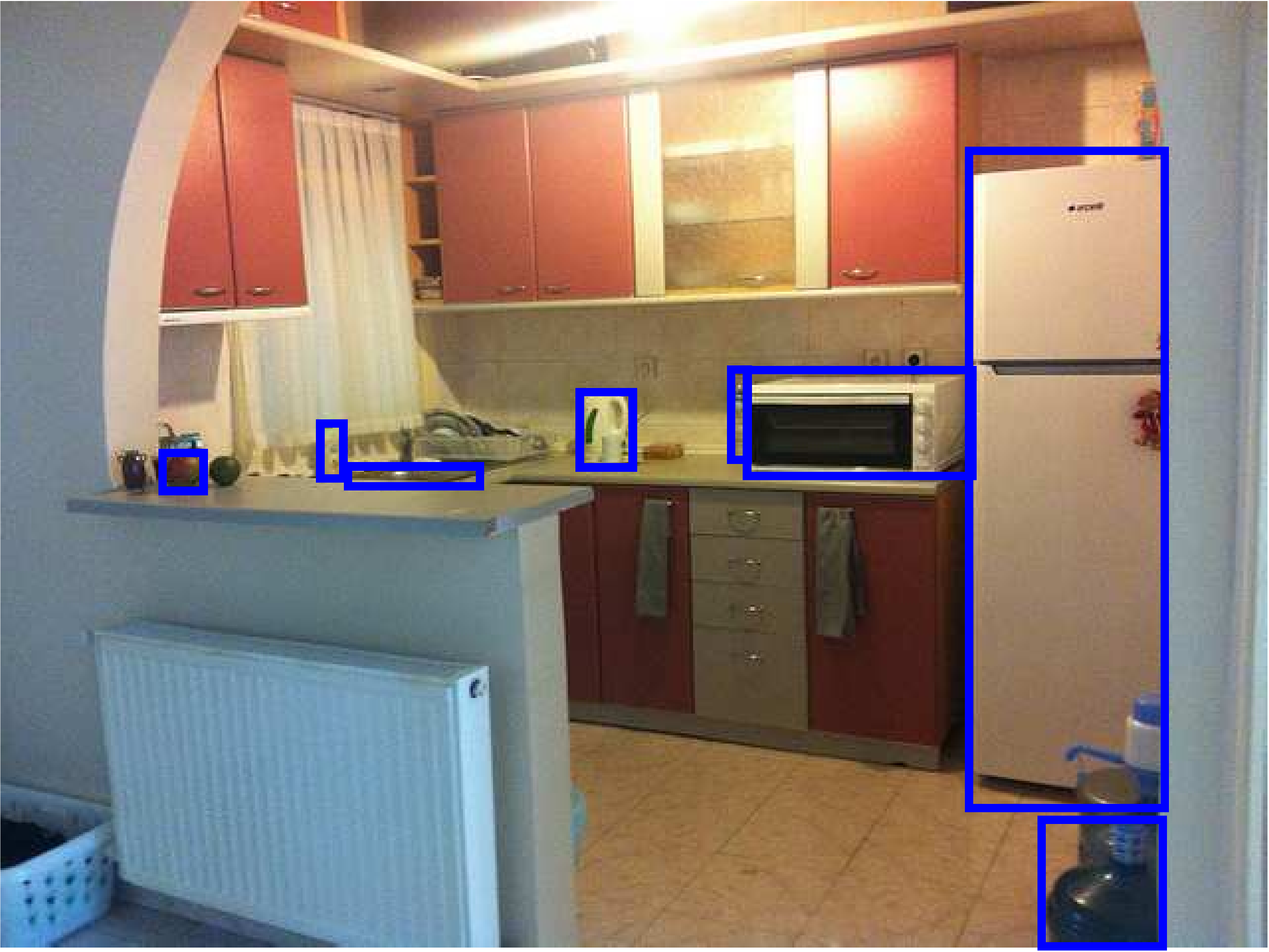} & \includegraphics[height=4cm]{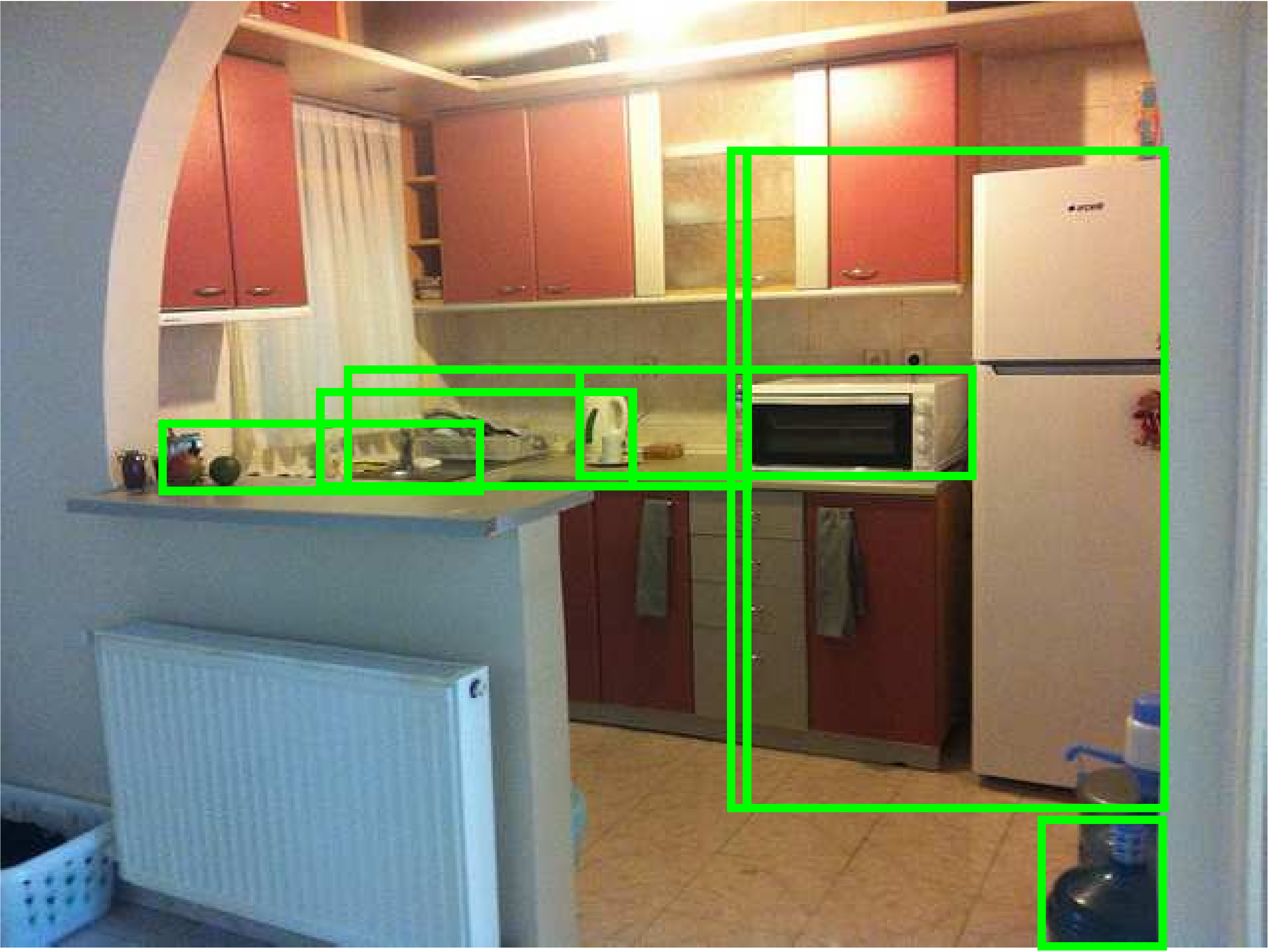} \\
			\includegraphics[width=5cm]{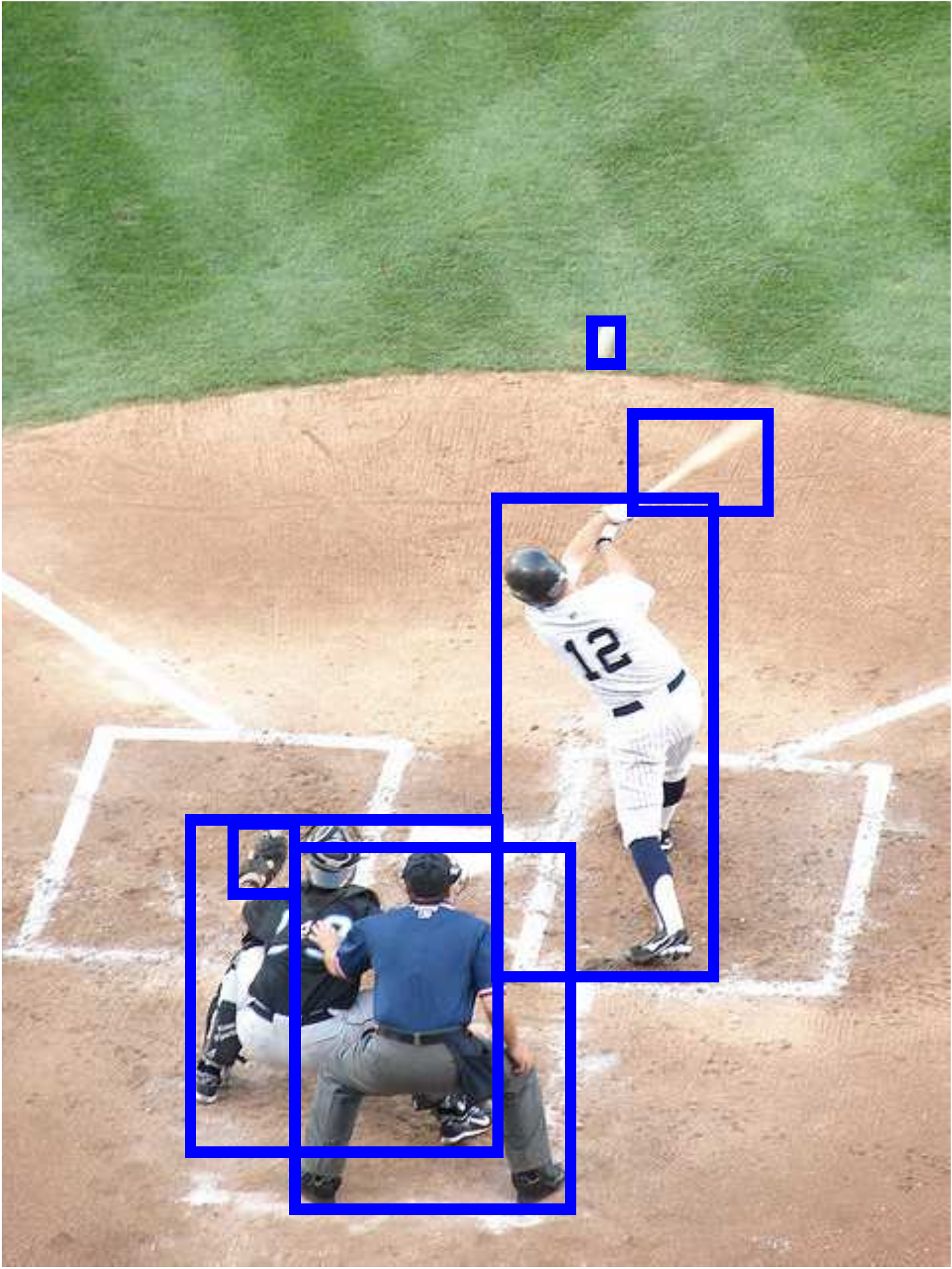} & \includegraphics[width=5cm]{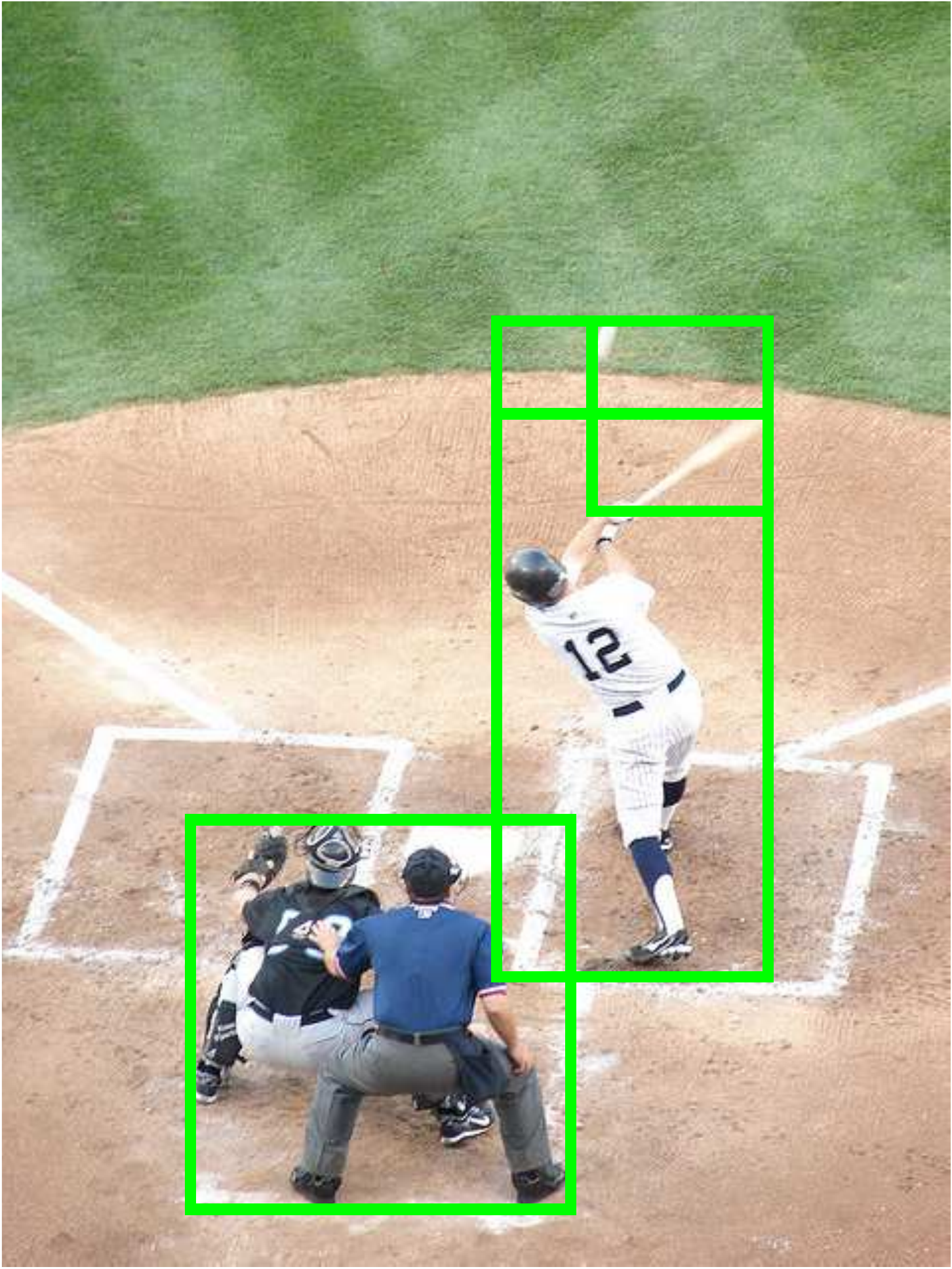} \\
			
	\end{tabular}}
	\caption{The comparison results between the object bounding box and generated multi-label region ground truth. Our generated regions contain multiple objects, for example, the generated regions contain the object of `person/baseball/baseball bat', 'cup/oven' and 'oven/fridge' altogether.}
	\label{fig:obj_reg_box_compare}
\end{figure}

\subsection{Evaluation Metrics}
We refer to the evaluation metrics used in \cite{gong2013deep,wang2016cnn} to compute precision and recall for the predicted labels. For each test image, we predict $k$ highest ranked labels and compare against the image ground-truth. The precision is the number of the correctly annotated labels divided by the number of predicted labels; the recall is the number of correctly annotated labels divided by the number of ground-truth labels. We compute \textit{overall precision \& recall (op \& or)} and \textit{per-class precision \& recall (cp \& cr)} based on the formulas blew. We also compute the \textit{mean average precision} (mAP) for the comparisons.
\begin{eqnarray}
op = \frac{\sum_{i=1}^{c}N_i^c}{\sum_{i=1}^{c}N_i^p} 
,\  or = \frac{\sum_{i=1}^{c}N_i^c}{\sum_{i=1}^{c}N_i^g} ,
\\cp = \frac{1}{c}\sum_{i=1}^{c}\frac{N_i^c}{N_i^p}
,\  cr = \frac{1}{c}\sum_{i=1}^{c}\frac{N_i^c}{N_i^g} ,
\end{eqnarray}

where $c$ is the number of total labels, $N_i^c$ is the number of images that are accurately labeled for $i_{th}$ label,$N_i^p$ is the number of the predicted images for $i_{th}$ label, and $N_i^g$ is the number of the ground truth images for $i_{th}$ label.

\subsection{Baseline Models}

To evaluate the effectiveness of our proposed RLSD model, we implement three baseline models for comparisons. The experimental results of these baseline models indicate the significance of our model's components.

\paragraph{Multi-label CNN} This is a standard CNN model without considering any label-dependencies. We use the VGGNet~\cite{simonyan2014very} pre-trained on the ImageNet for parameters initialization, and fine-tune the model on the benchmark datasets. The dimension of the last fully-connected layer is changed to the class number of each dataset and the element-wise logistic loss is applied. The learning rates of the last two fully-connected layers are initialized as 0.001 and 0.05 respectively, the rest of the layers are fixed. We run around 45 epochs in total until the model leads to convergence and decease the learning rate to one tenth every 15 epochs, the weight decay is 0.0005 and momentum is 0.9.

\paragraph{CNN+LSTM} This is a model with the same configurations as our proposed RLSD model, except that it only considers the label dependencies at the global level. We use the last fully-connected layer of the pre-trained VGGNet \cite{simonyan2014very} as the global image representation. Then we feed the image feature into the LSTM to train a multi-label classification model. This baseline is similar to the CNN-RNN model proposed in \cite{simonyan2014very}, except the image feature is only fed into the LSTM once at the first time step. The dimensions of label embedding and LSTM layer are the same as our proposed model: 64 and 512 respectively.

\paragraph{MCG-CNN+LSTM} This is another baseline model, which is used to verify the effectiveness of our localization of multi-label regions. In this model, we use the exactly same configurations as our proposed model in Sec.~\ref{sec:model} except that the region proposal network is replaced by an object proposal tool: the Multiscale Combinatorial Grouping (MCG) \cite{PABMM2015}. The pre-extracted top-256 object proposals (based on the object proposal confidence score) for each image are sent into our region-based LSTM for training and testing. As shown in Fig.~\ref{fig:region_proposals}, our region-based localization layer can generate the proposals with multiple labels.

\subsection{Results on the VOC PASCAL 2007}

VOC PASCAL 2007 \cite{Everingham15} is a benchmark dataset for image classification, segmentation and detection. In VOC 2007 dataset, there are 9,963 images in total, of which 5,011 and 4,952 images for training/validation and testing respectively. 20 common objects are annotated in this dataset.

Tab.~\ref{voc2007} shows the comparisons to the state-of-the-art methods including the HCP-1000C/2000C \cite{single2multi} and CNN-RNN \cite{wang2016cnn}, along with the baseline methods on VOC 2007 dataset. The HCP-1000C uses image region as the input and fine-tunes the CNN pre-trained on ImageNet 1000 classes. The HCP-2000C augments the ILSVRC-2012 training set with additional 1,000 ImageNet classes to pre-train the model. The CNN-RNN model \cite{wang2016cnn}, as we mentioned before, explores the image multi-label dependencies at the global level. Our proposed RLSD model uses the pre-trained region proposal network and fixes the localization layer during the training, while the upper bound mode RLSD+ft-RPN fine-tunes the localization layer with the region-based bounding-boxes as the guidance. The region bounding-boxes generation details have been discussed in the Sec.~\ref{details}. From Tab.~\ref{voc2007}, we can see that our RLSD model outperforms those state-of-the-art methods and baseline models in most classes. It achieves the 87.3\% mAP, while the CNN-RNN model \cite{wang2016cnn} is 84\% and our own CNN+LSTM baseline is 83.5\%. This proves the regional semantic dependencies are more advanced than only considering the label dependencies at the global level. The gap between the Multi-CNN and our RLSD~(4.3\%) is larger than the gap between Multi-CNN and CNN+LSTM~(0.5\%), which also stresses this point.  Although the MCG-CNN+LSTM model uses the regional information, the region proposal from the MCG \cite{PABMM2015} normally only contains a single object, which means the label dependencies are invalid on those proposals anymore. In our RLSD model, the region-based localization layer can localize the regions that contain multiple highly-dependent labels. Therefore, our RLSD model also outperforms the MCG-CNN+LSTM model with a big margin, 87.3\% VS. 84.9\%, which is only 1.2\% lower than the upper bound model RLSD+ft-RPN.


\begin{table}[t]
	\centering
	\resizebox{\linewidth}{!}{
		\rowcolors{2}{gray!25}{} 
		\begin{tabular}{cccc|ccc}
			\Xhline{2\arrayrulewidth}
			& C-P  & C-R  & C-F1 & O-P  & O-R  & O-F1 \\ \Xhline{2\arrayrulewidth}
			Multi-CNN   & 47.5 & 85.2 & 61.0 & 46.4 & 90.2 & 61.3 \\
			CNN+LSTM    & 46.7 & 88.9 & 61.2 & 47.0 & 91.5 & 62.1 \\
			MCG-CNN+LSTM    & 48.6 & 88.7 & 62.8 & 46.9 & 91.4 & 62.0 \\ 
			\hline
			RLSD        & 50.5 & 90.6 & 64.9 & 47.5 & 92.4 & 62.7 \\
			RLSD+ft-RPN & \textbf{51.6} & \textbf{90.7} & \textbf{65.8} & \textbf{47.6} & \textbf{92.7} & \textbf{62.9} \\ \Xhline{2\arrayrulewidth}
	\end{tabular}}
	\vspace{-1pt}
	\caption{Comparisons on the PASCAL VOC 2007 for different evaluation metrics, $k=3$.}
	\label{PR-voc2007}
	\vspace{-13pt}	
\end{table}

By introducing the region information into our model, it surpasses the current methods in a big gap for predicting small objects and visual concepts such as `bottle', `TV' and `chair' etc. Take the `bottle' as an example, we achieve 10\% higher than CNN-RNN model. 
Tab.~\ref{PR-voc2007} compares the per-class precision \& recall (cp \& cr), overall precision \& recall (op \& or), and F1 score between our proposed models and the baseline models. Our RLSD outperforms the baselines on all the evaluation metrics, and approaches the RLSD+ft-RPN results. Particularly, our model has achieved a much higher precision, no matter for per-class or overall evaluations, and is very close to the upper bound results.

\subsection{Results on the Microsoft COCO}
Microsoft COCO dataset \cite{lin2014microsoft} is a large scale benchmark dataset for several vision tasks. There are in total 123,287 images for training and validation with 80 object concepts annotated. We use all the annotated object labels in an image as the multi-label ground-truth, and employ its training set as the training data and validation set as the test data. After removing the images without annotation, we have 82,081 images for training and 40,137 images for testing. We obtain the semantic dependencies of these labels by computing their co-occurrence rates and form them as a matrix. We have found that there are strong dependencies in its label set, e.g. `keyboard' and `computer' always appear together.

%
%
%


\begin{figure*}[t]
	\vspace{-5pt}
	\centering
	\resizebox{\linewidth}{!}{
		\begin{tabular}{c c c}
			\includegraphics[width=0.7\linewidth]{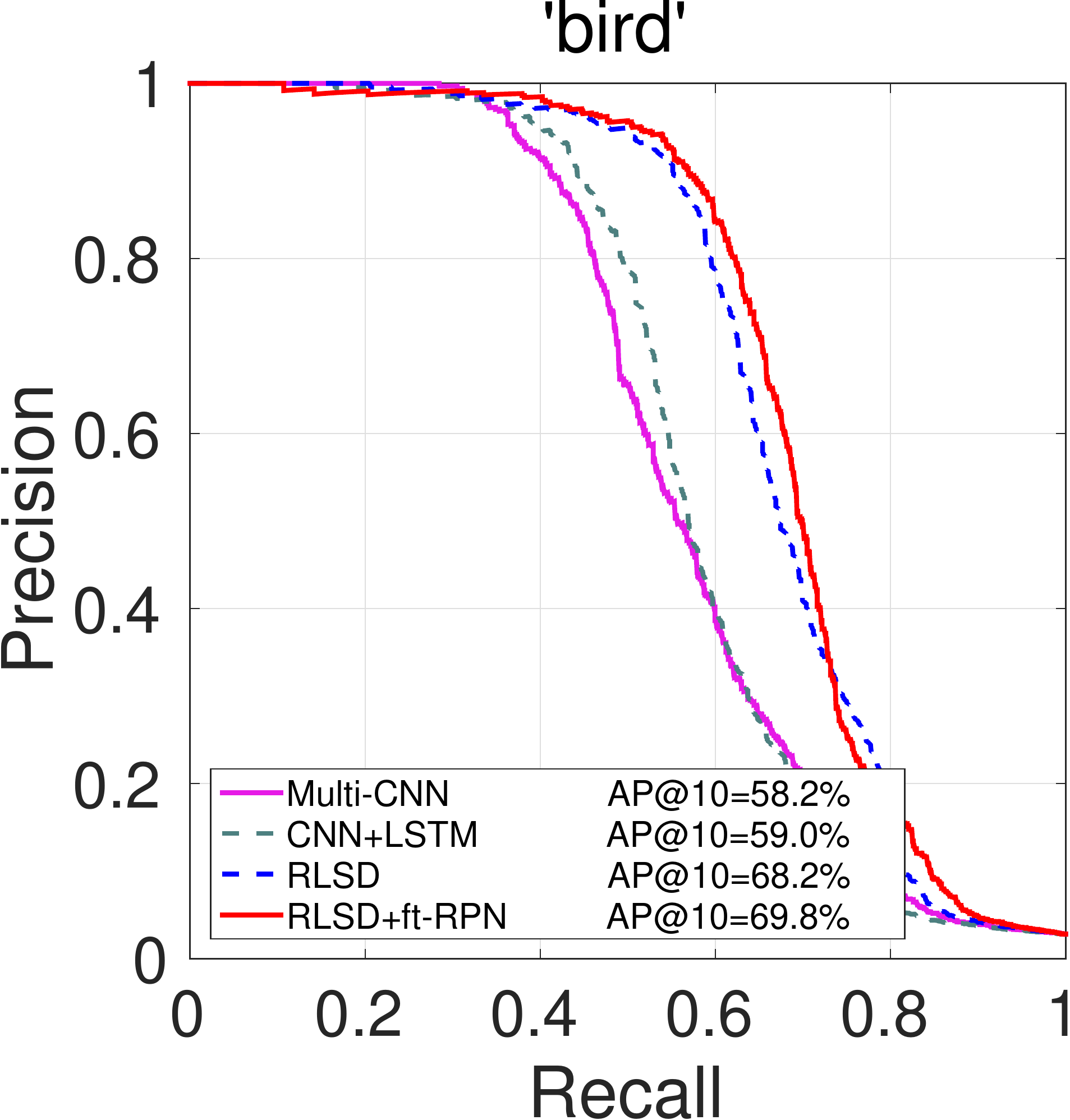} & \includegraphics[width=0.7\linewidth]{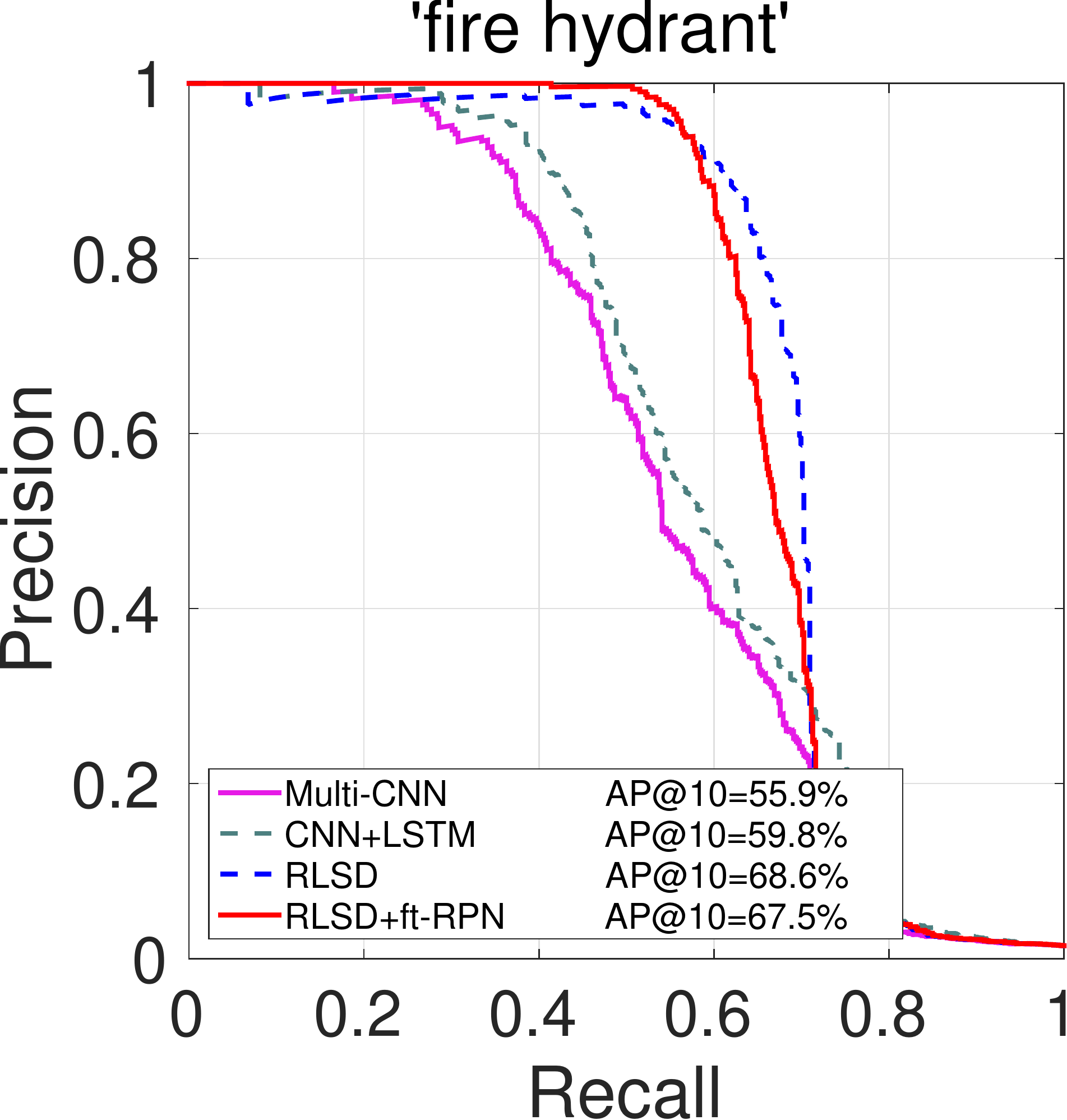}&
			\includegraphics[width=0.7\linewidth]{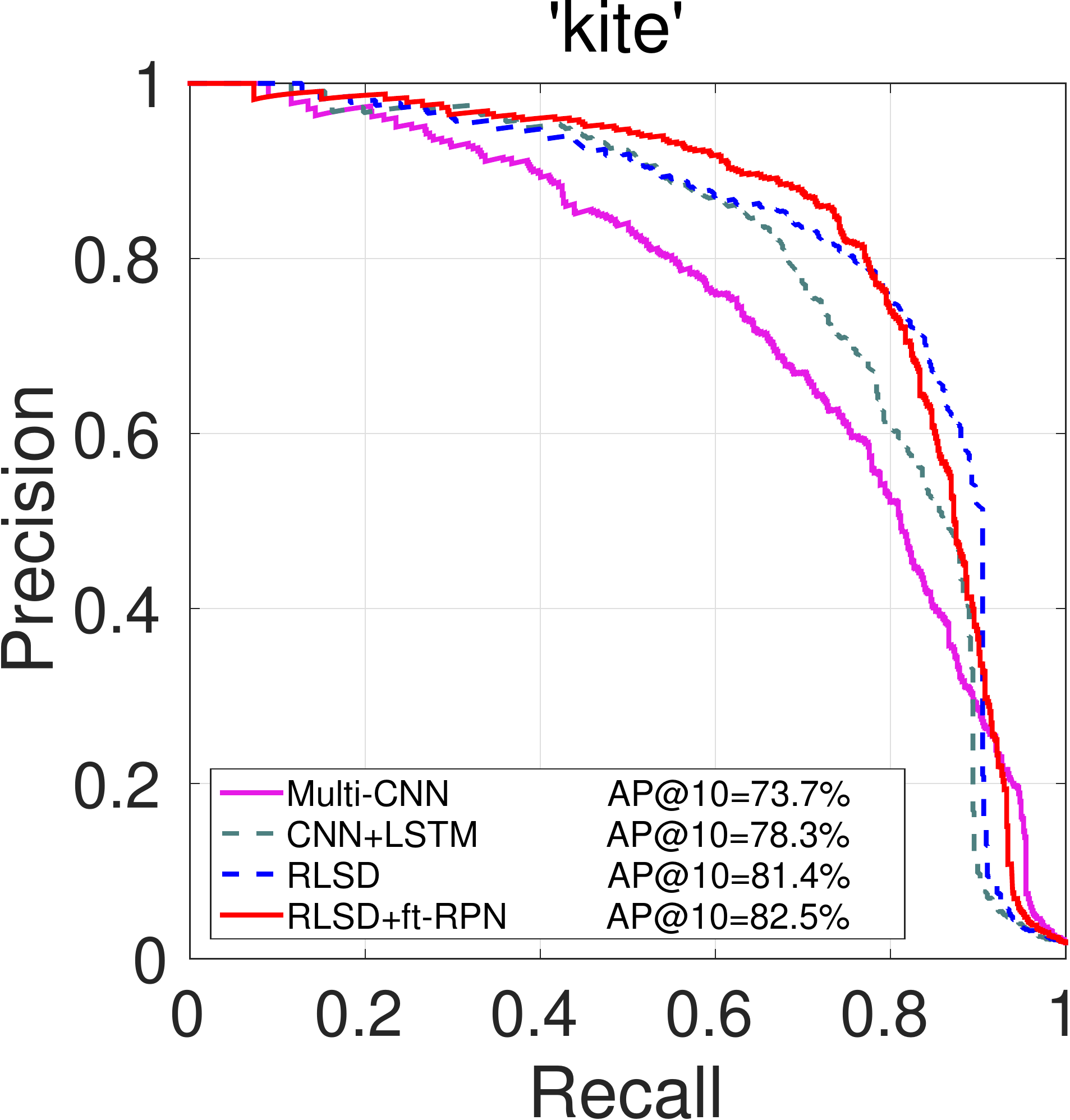}\\
	\end{tabular}}
	\caption{Precision-Recall curves for the `bird', `fire hydrant' and `kite' classes in the MS COCO dataset, for our RLSD models and the baseline models. The average precision @ 10 are also given in the figure.
	}
	\label{fig:prcurve}
\end{figure*}

\begin{figure}[t]
	\centering
	\includegraphics[width=7.3cm,height=6.4cm]{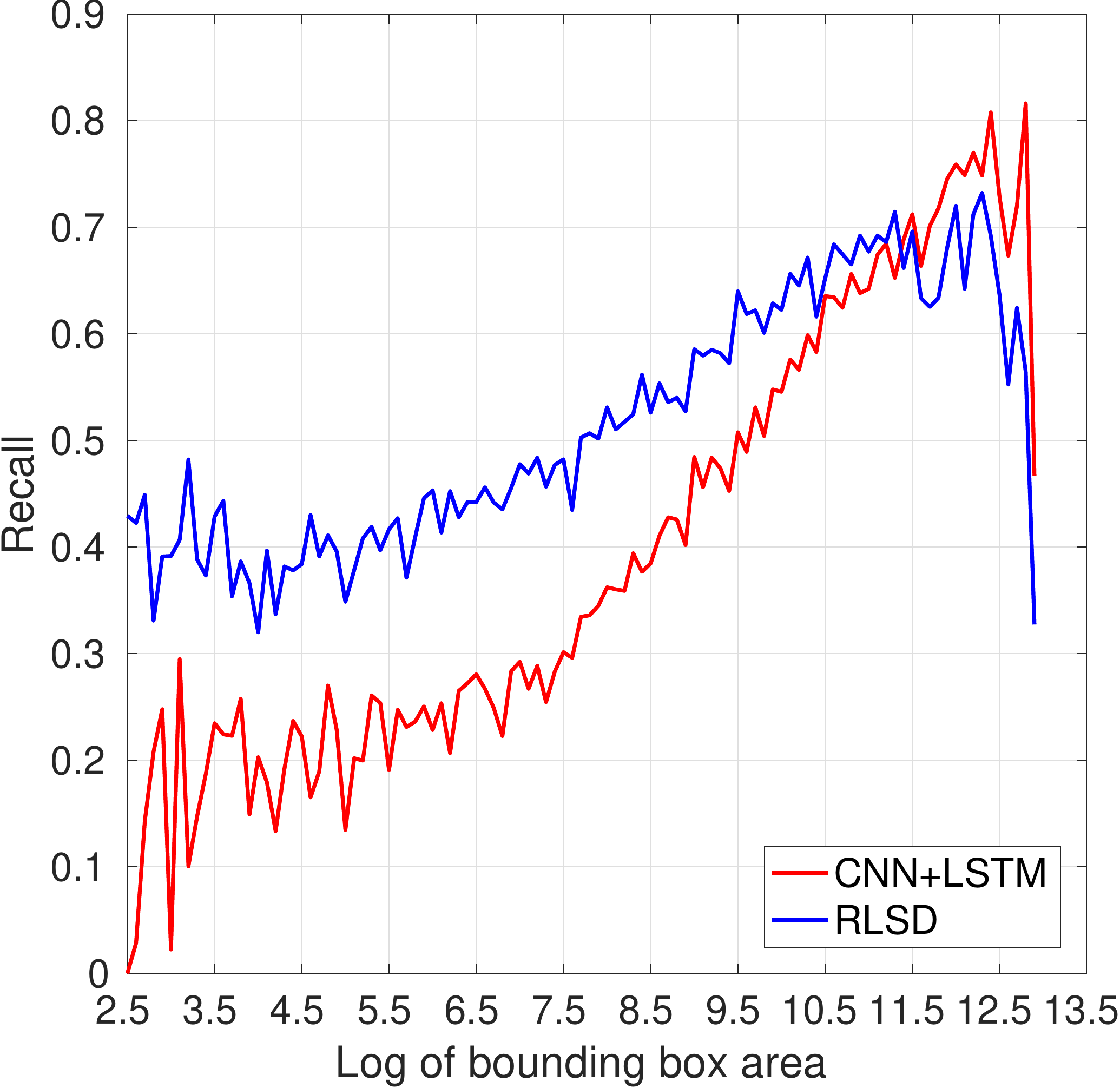}
	\caption{The relationship between recall and bounding-box area
		on Microsoft COCO dataset.}
	\label{fig:bdbox_recall}
\end{figure}

\begin{figure*}[t!]
	\resizebox{\linewidth}{!}{
		\begin{tabular}{ccccc}
			\multicolumn{2}{c}{
				\includegraphics[height=3cm]{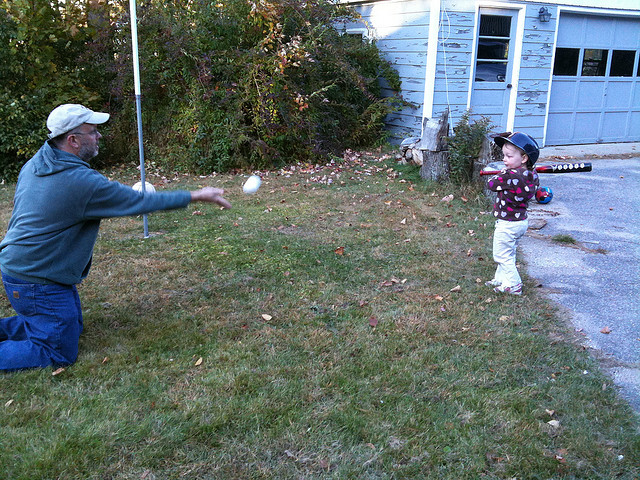}}&
			\includegraphics[height=3cm]{./img/coco_finalpred/17708.jpg}&
			\includegraphics[height=3cm]{./img/coco_finalpred/35269.jpg}&
			\includegraphics[height=3cm]{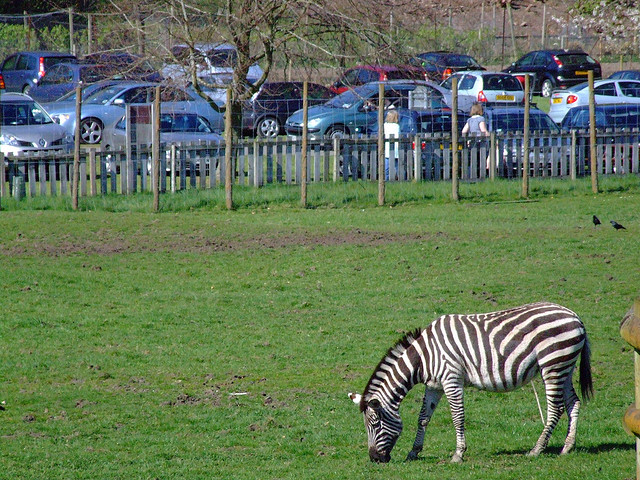}\\
			\multicolumn{2}{c}{(A)}&(B)&(C)&(D)\\
			\hline\noalign{\smallskip}\noalign{\smallskip}
			Ground Truth:&
			person, sports ball, baseball bat&
			person, boat, bird&
			bottle, cup, fork, knife, sandwich, dining table&
			person, car, bird, zebra\\
			Multi-CNN:&
			\color{red}{person}&
			\color{red}{person}&
			\color{red}{dining table}&
			\color{red}{zebra}\\
			CNN+LSTM:&
			\color{red}{person, baseball bat, frisbee}&
			\color{red}{person, boat, dog}&
			\color{red}{cup, fork, knife, pizza, sandwich, dining table}&
			\color{red}{zebra, car}\\
			RLSD:&
			\color{green}{person, \textbf{\underline{sports ball}}, baseball bat}&
			\color{green}{person, boat, \textbf{\underline{bird}}}&
			\color{green}{\textbf{\underline{bottle}}, cup, fork, knife, sandwich, dining table}&
			\color{green}{person, car, \textbf{\underline{bird}}, zebra}\\
			\noalign{\smallskip}\hline\noalign{\smallskip}\noalign{\smallskip}\\
			
			\multicolumn{2}{c}{
				\includegraphics[height=3cm]{./img/coco_finalpred/121632.jpg}}&
			\includegraphics[height=3cm]{./img/coco_finalpred/135756.jpg}&
			\includegraphics[height=3cm]{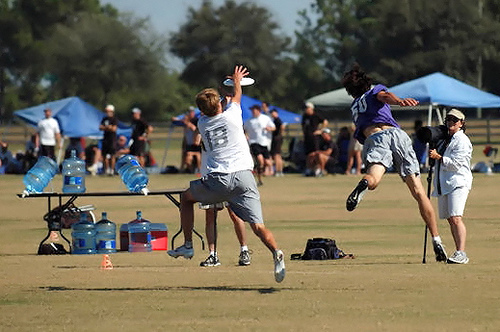}&
			\includegraphics[height=3cm]{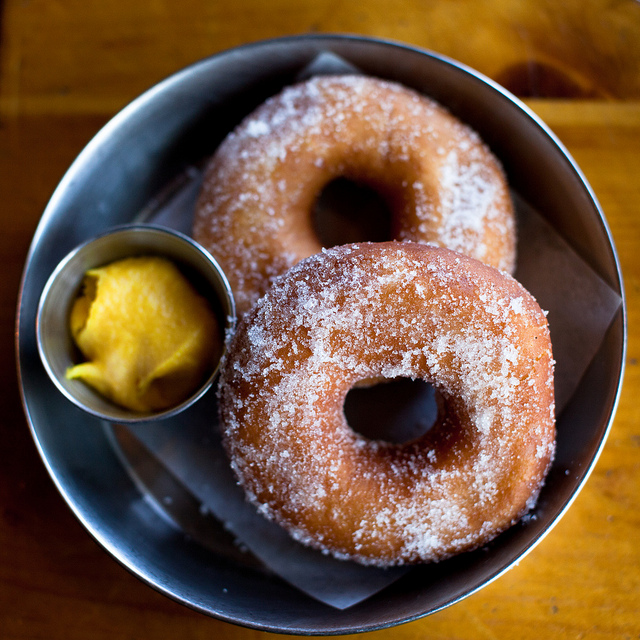}\\
			\multicolumn{2}{c}{(E)}&(F)&(G)&(H)\\
			\hline\noalign{\smallskip}\noalign{\smallskip}
			Ground Truth:&
			cup, fork, knife, pizza&
			person, horse, bottle, car&
			person, frisbee, bottle&
			bowl, donut, dining table\\
			Multi-CNN:&
			\color{red}{pizza, dining table}&
			\color{red}{person}&
			\color{red}{person}&
			\color{red}{donut}\\
			CNN+LSTM:&
			\color{red}{fork, pizza, dining table}&
			\color{red}{person, horse}&
			\color{red}{person, sports ball, backpack, tennis racket}&
			\color{red}{dount dining table}\\
			RLSD:&
			\color{green}{\textbf{\underline{cup}}, fork, knife, pizza}&
			\color{green}{person, horse, \textbf{\underline{bottle}}, car}&
			\color{green}{person, \textbf{\underline{frisbee}}, \textbf{\underline{bottle}}, umbrella}&
			\color{green}{bowl, donut, dining table}\\
			\noalign{\smallskip}\hline\noalign{\smallskip}\noalign{\smallskip}\\
			
			\multicolumn{2}{c}{
				\includegraphics[height=3cm]{./img/coco_finalpred/282143.jpg}}&
			\includegraphics[height=3cm]{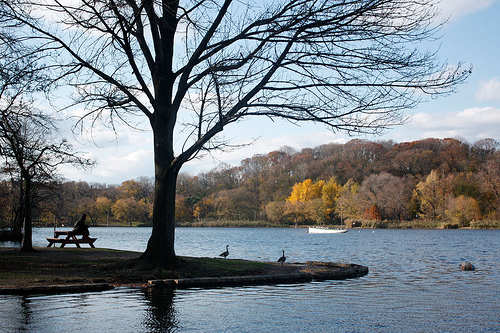}&
			\includegraphics[height=3cm]{./img/coco_finalpred/392229.jpg}&
			\includegraphics[height=3cm]{./img/coco_finalpred/525666.jpg}\\
			\multicolumn{2}{c}{(I)}&(J)&(K)&(L)\\
			\hline\noalign{\smallskip}\noalign{\smallskip}
			Ground Truth:&
			boat, cup, fork, knife, pizza, dining table&
			person, boat, bench, bird&
			wine glass, cup, fork, knife, pizza&
			person, boat, sports ball, surf board\\
			Multi-CNN:&
			\color{red}{pizza, dining table}&
			\color{red}{boat}&
			\color{red}{pizza, dining table, fork, knife}&
			\color{red}{person}\\
			CNN+LSTM:&
			\color{red}{fork, pizza, dining table, cup, bottle}&
			\color{red}{person, boat}&
			\color{red}{pizza, dining table, fork, knife}&
			\color{red}{person, surf board}\\
			RLSD:&
			\color{green}{boat, cup, fork, \textbf{\underline{knife}}, pizza, dining table}&
			\color{green}{person, boat, bench, \textbf{\underline{bird}}}&
			\color{green}{\textbf{\underline{wine glass}}, cup, fork, knife, pizza}&
			\color{green}{person, boat, \textbf{\underline{sports ball}}, surf board}\\
			\noalign{\smallskip}\hline\noalign{\smallskip}\noalign{\smallskip}\\
	\end{tabular}}
	\caption{Some example multi-label classification results from the MS COCO.}
	\label{results_examples} 
\end{figure*}


\begin{table}[t!]
	\centering
	\resizebox{\linewidth}{!}{
		\rowcolors{2}{gray!25}{} 
		\begin{tabular}{cccc|ccc|cc}
			\Xhline{2\arrayrulewidth}
			& C-P  & C-R  & C-F1 & O-P  & O-R  & O-F1 & mAP&mAP@10   \\ \Xhline{2\arrayrulewidth}
			WARP \cite{gong2013deep}        & 59.3 & 52.5 & 55.7 & 59.8 & 61.4 & 60.7 & -&49.2      \\
			CNN-RNN \cite{wang2016cnn}    & 66.0 & 55.6 & 60.4 & 69.2 & \textbf{66.4} & \textbf{67.8} & -&61.2     \\ \hline
			Multi-CNN   & 54.8 & 51.4 & 53.1 & 56.7 & 58.6 & 57.6 & 60.4&57.8    \\
			CNN+LSTM    & 62.1 & 51.2 & 56.1 & 68.1 & 56.6 & 61.8 & 61.8&59.2    \\
			MCG-CNN+LSTM    & 64.2 & 53.1 & 58.1 & 61.3 & 59.3 & 61.3 & 64.4&62.4    \\ \hline
			RLSD        & \textbf{67.7} & 56.4 & 61.5 & \textbf{70.5} & 59.9 & 64.8 & 67.4&65.9    \\
			RLSD+ft-RPN & 67.6 & \textbf{57.2} & \textbf{62.0} & 70.1 & 63.4 & 66.5 &\textbf{68.2} &\textbf{66.7}   \\ \Xhline{2\arrayrulewidth}
	\end{tabular}}
	\caption{Comparisons on the MS-COCO dataset for $k=3$. mAP@10 measures are additionally computed for comparison.}
	\label{COCO}	
\end{table}

We compare our RLSD model with the WARP \cite{gong2013deep} model and CNN-RNN \cite{wang2016cnn} model in Tab.~\ref{COCO}. Our RLSD model outperforms the CNN-RNN \cite{wang2016cnn} on the evaluation metrics of per-class precision \& recall, and overall precision. Although we have a lower overall recall, because our model may predict less than $k$ labels for an image (a threshold $t=0.5$ is set on the prediction score to decide whether the label should be output). We have higher mAP and mAP@10 measures than the state-of-the-art methods. We also outperform the MCG-CNN+LSTM model, which only uses the MCG object proposals as the input. The RLSD+ft-RPN results shows our model is still close to the upper bound even without the RPN fine-tuning, which indicates the generality of our model in the real world.


Fig.~\ref{fig:prcurve} shows the precision-recall curves of three selected classes (bird, fire hydrant and kite) in the MS COCO dataset, these three kinds of objects are visually small in the images of this dataset. The PR curves illustrate that our RLSD model significantly surpasses the baseline models and is especially good at predicting these small objects.

Some qualitative results are shown in Fig.~\ref{results_examples}. Small sized objects are noted with underline. As we can see, small objects like `sports ball' in image $(A)$, `bird' in image $(B)$, and `bottle' in image $(F)$ etc. are only predicted by our RLSD model.

Fig.~\ref{fig:bdbox_recall} analyzes the recall along with the object ground-truth bounding-box size on the MS COCO dataset, which shows that our model achieves much higher recall than the CNN+LSTM model on those labels that corresponding bounding-boxes are smaller. This means that our RLSD model is more sensitive to `small' objects. Both recall curves start to fall when the object is so large that it almost fills the whole image, similar case is also observed in \cite{wang2016cnn}.


\subsection{Results on the NUS-WIDE}
NUS-WIDE \cite{nus-wide-civr09} is a web image dataset associated with user tags. 
The whole dataset contains 269,648 images and 1,000 tags. These images are further manually labeled into 81 concepts. After removing the no-annotated images, the training set and test set contains 125,449 and 83,898 images respectively.

Tab.~\ref{nus} shows the comparison results on the NUS-WIDE dataset on 81 concepts. We compare the proposed RLSD model with previous mentioned methods including metric learning \cite{li2015distributed}, multi-edge graph \cite{liu2010unified}, K nearest neighbor~\cite{nus-wide-civr09}, softmax prediction, WARP method \cite{gong2013deep} and the latest state-of-the-art method CNN-RNN \cite{wang2016cnn}. Same as~\cite{wang2016cnn}, we do not fine-tune the CNN image representation. Our proposed RLSD model outperforms these methods in a large margin. Specifically, we achieve 4\% higher precision than the CNN-RNN. Because the bounding-boxes are not provided in the NUS-WIDE dataset, there is no upper bound for this dataset.

\begin{table}[t]
	\vspace{2pt}
	\centering
	\resizebox{\linewidth}{!}{
		\rowcolors{2}{gray!25}{} 
		\begin{tabular}{cccc|ccc|c}
			\Xhline{2\arrayrulewidth}
			& C-P  & C-R  & C-F1 & O-P  & O-R  & O-F1 & mAP   \\ \Xhline{2\arrayrulewidth}
			Metric Learning \cite{li2015distributed}        & - & - & - & - & - & 21.3 & -      \\
			Multi-edge graph \cite{liu2010unified}      & - & - & - & 35.0 & 37.0 & 36.0 & -      \\
			KNN \cite{nus-wide-civr09}        & 32.6 & 19.3 & 24.3 & 42.9 & 53.4 & 47.6 & -      \\
			softmax \cite{wang2016cnn}     & 31.7 & 31.2 & 31.4 & 47.8 & 59.5 & 53.0 & -      \\
			WARP \cite{gong2013deep}       & 31.7 & 35.6 & 33.5 & 48.6 & 60.5 & 53.9 & -      \\
			CNN-RNN \cite{wang2016cnn}     & 40.5 & 30.4 & 34.7 & 49.9 & 61.7 & 55.2 & -     \\ \hline
			Multi-CNN   & 40.2 & 42.8 & 41.5 & 53.4 & 66.4 & 59.2 & 47.1    \\
			CNN+LSTM    & 42.2 & 45.8 & 43.9 & 53.4 & 66.5 & 59.3 & 49.9   \\
			MCG-CNN+LSTM    & 44.3 & 48.1 & 46.1 & 54.1 & 67.2 & 59.9 & 52.4    \\ \hline
			RLSD        & \textbf{44.4} & \textbf{49.6} & \textbf{46.9} & \textbf{54.4} & \textbf{67.6} & \textbf{60.3} & \textbf{54.1}    \\ \Xhline{2\arrayrulewidth}
	\end{tabular}}
	\caption{Comparisons on the NUS-WIDE on 81 concepts for $k=3$.}
	\label{nus}
\end{table}

	\section{Conclusion}
	Multi-label image classification is an important problem on multimedia, because it is not only more challenging than the single-label image classification, but also closer to the real-world applications. In this paper, we propose a Regional Latent Semantic Dependencies (RLSD) model to address this problem. As its name suggests, this proposed model can capture the label dependencies at the regional level. Experimental results on several benchmark datasets show that the proposed RLSD model consistently achieves the superior overall performance to the state-of-the-art approaches, especially for predicting small objects and visual concepts occurred in the image.
	
	In the future, we will investigate to localize the multi-label regions in an unsupervised manner. The attention mechanism can be valued in this situation since it can be used to model the spatial relationship between labels. We will combine the attention mechanism with our proposed regional latent semantic dependencies model in our future work.

	%
	%

	\ifCLASSOPTIONcaptionsoff
	\newpage
	\fi

	{\small
		\bibliographystyle{IEEEtran}
		\bibliography{mybib}
	}

\end{document}